\newif\ifarxiv
\arxivtrue 

\ifarxiv
    \documentclass{article}
\else
    \documentclass[submission]{style/hld2026}
\fi
\usepackage{style/macros}
\usepackage[T1]{fontenc}
\usepackage[utf8]{inputenc}
\usepackage{lmodern}
\usepackage{libertine}
\usepackage{hyphenat}
\usepackage{natbib}
\usepackage{subcaption}
\usepackage{caption}

\usepackage{libertine}

\ifarxiv
    \usepackage[letterpaper, margin=1in]{geometry}
    \hypersetup{colorlinks, linkcolor=blue, citecolor=blue, urlcolor=magenta}
\fi

\hypersetup{
    pdftitle={A "feature ODE" describing the learning behavior of shallow MLPs on simple functions},
    pdfauthor={Joseph Turnbull, Berkan Ottlik, Jamie Simon}
}

\begin{document}

\ifarxiv

    \title{A ``feature ODE'' describing the learning behavior of shallow MLPs on simple functions}
    \author{%
    Joseph Turnbull$^{1,*}$
    \qquad
    Berkan Ottlik$^{2,*}$
    \qquad
    Jamie Simon$^{1,3}$
    \\[1ex]
    {\normalsize
    $^1$UC Berkeley
    \qquad
    $^2$University of Pennsylvania
    \qquad
    $^3$Imbue}
    \\[0.5ex]
    {\small
    \texttt{joeyturnbull@berkeley.edu},
    \
    \texttt{ottlik@seas.upenn.edu},
    \
    \texttt{jsi@berkeley.edu}}
    }
    \date{}
\fi

\maketitle

\begingroup
    \let\thefootnote\relax
    \footnotemark\footnotetext{\hspace{-6pt} $^\star$Joint primary authorship.}
    \footnotemark\footnotetext{\hspace{-2pt} Code: \url{https://github.com/berkott/feature-ode}}
\endgroup

\begin{abstract}

We study the gradient flow training dynamics of two-layer multi-layer perceptrons (MLPs) on anisotropic Gaussian data.
Even in this simple setting, solving the dynamics in full is notoriously difficult, so we try something different.
We propose a ``feature ODE'' that might be intuitively expected to describe the learning dynamics of a shallow MLP.
We find that MLPs trained to learn simple target functions (low-order Hermite polynomials and staircase functions) actually roughly follow the loss trajectories from this feature ODE.
This agreement is a surprise because of the size of the simplification and the fact that it does not come from a rigorous chain of approximations.
We discuss this surprise and speculate on its reasons and uses.

\end{abstract}

\begin{figure}[h]
    \centering
    \includegraphics[width=\textwidth]{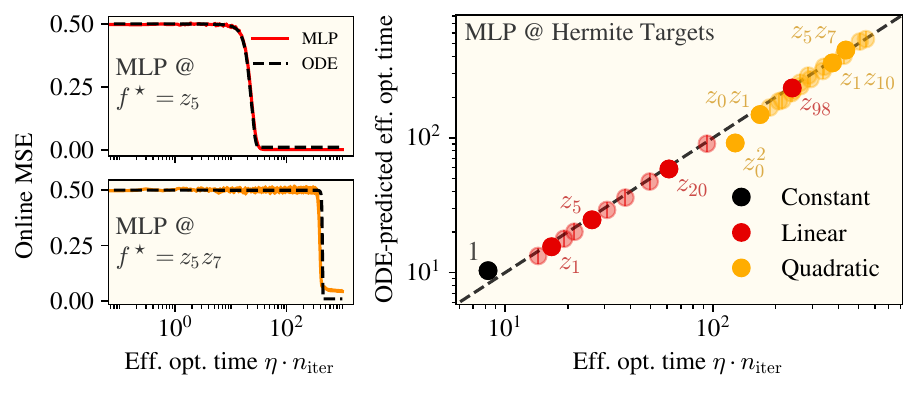}
    \caption{
    \textbf{We suggest a simple ODE of scalar variables whose solution predicts loss trajectories for shallow MLPs learning simple functions of anisotropic data.} These plots treat shallow MLPs with shifted SiLU activation functions ($\sigma(x)=\text{SiLU}(x+0.5)$) learning various Hermite polynomial target functions data sampled $\vx \sim \N(0, \mGamma)$ with $\mGamma = \text{diag}(\gamma_i)$, $\gamma_i \propto (i+6)^{-1.8}$. We write $z_i\equiv\vx_i/\sqrt{\gamma_i}$ for normalized coordinates. \textbf{Left.} MLP loss trajectories closely match the ODE for example linear and quadratic target functions. \textbf{Right.} Effective optimization times $\eta \cdot n_\text{iter}$ required for learning to complete are well-predicted by this ODE across many such target functions. For quadratic terms of the form $\text{He}_2(z_i)$, the leading term is indicated in the plots ($z_i^2$). See \cref{app:experimental_setup} for experimental details.
    }
    \label{fig:main_figure}
\end{figure}

\section{Introduction}
\label{sec:intro}

How do shallow MLPs learn simple functions of anistropic Gaussian data?
Many researchers have studied these training dynamics, and we have learned a lot (see \cref{app:related-works}).
%
One definite conclusion has emerged: the problem is hard — surprisingly hard, for what should be a simple toy problem in deep learning theory.
 
%
%
%
What is the source of the difficulty?
Taking a look at the functional form of a shallow MLP,
\begin{equation}
    f_\text{MLP}(\vx) = \langle \vw_{\text{out}}, \sigma\left( \mW_\text{in} \vx \right) \rangle,    
    \label{eqn:mlp_functional_form}
\end{equation}
the main annoyance is the elementwise nonlinearity $\sigma(\cdot)$.
Elementwise nonlinearities do not play well with our nicest mathematical tools:
try to exactly solve the dynamics of a model parameterized as \cref{eqn:mlp_functional_form} using e.g. regular matrix ODEs, and the elementwise nonlinearity (which is rather unnatural from a linear-algebraic point of view) gets in the way.
Rather than develop a new tool for treating the exact dynamics of a shallow MLP, we propose something different altogether.
We design our system to be a solvable model of the shallow MLP that keeps two of the MLP's key traits.
Firstly, there should be two trainable linear transformations with a (fixed) nonlinear operation in between, and secondly, the model class needs to represent any nonlinear function.
We get both of these traits in the following ``deep polynomial regression'' (DPR) model:
\begin{equation}
    f_\text{DPR}(\vx) = \langle \mB, \vpsi_\vc(\mA \vx) \rangle,
    \label{eqn:dpr_functional_form}
\end{equation}
where $\mA$ is a square matrix, $\mB$ is a readout vector, and $\vpsi_\vc$ is the following \textit{polynomial featurization} operation:
\newlength{\prismarraysep}
\setlength{\prismarraysep}{1.5em}
\begin{equation}
    \vpsi_{\vc}(\vh)
    := \text{vec}\left(
        \left[
        \begin{array}{@{}c@{\hspace{\prismarraysep}}c@{\hspace{\prismarraysep}}c@{\hspace{\prismarraysep}}c@{\hspace{\prismarraysep}}c@{}}
            \sqrt{c_0}
            & \sqrt{c_1}\,\vh
            & \frac{\sqrt{c_2}}{\sqrt{2!}}\,\vh\vh^\top
            & \frac{\sqrt{c_3}}{\sqrt{3!}}\,\vh^{\otimes 3}
            & \cdots
        \end{array}
        \right]
    \right),
\end{equation}
with $\vc = (c_\ell)$ a sequence of constants.

\begin{figure}[ht]
    \centering
    \includegraphics[width=0.8\textwidth]{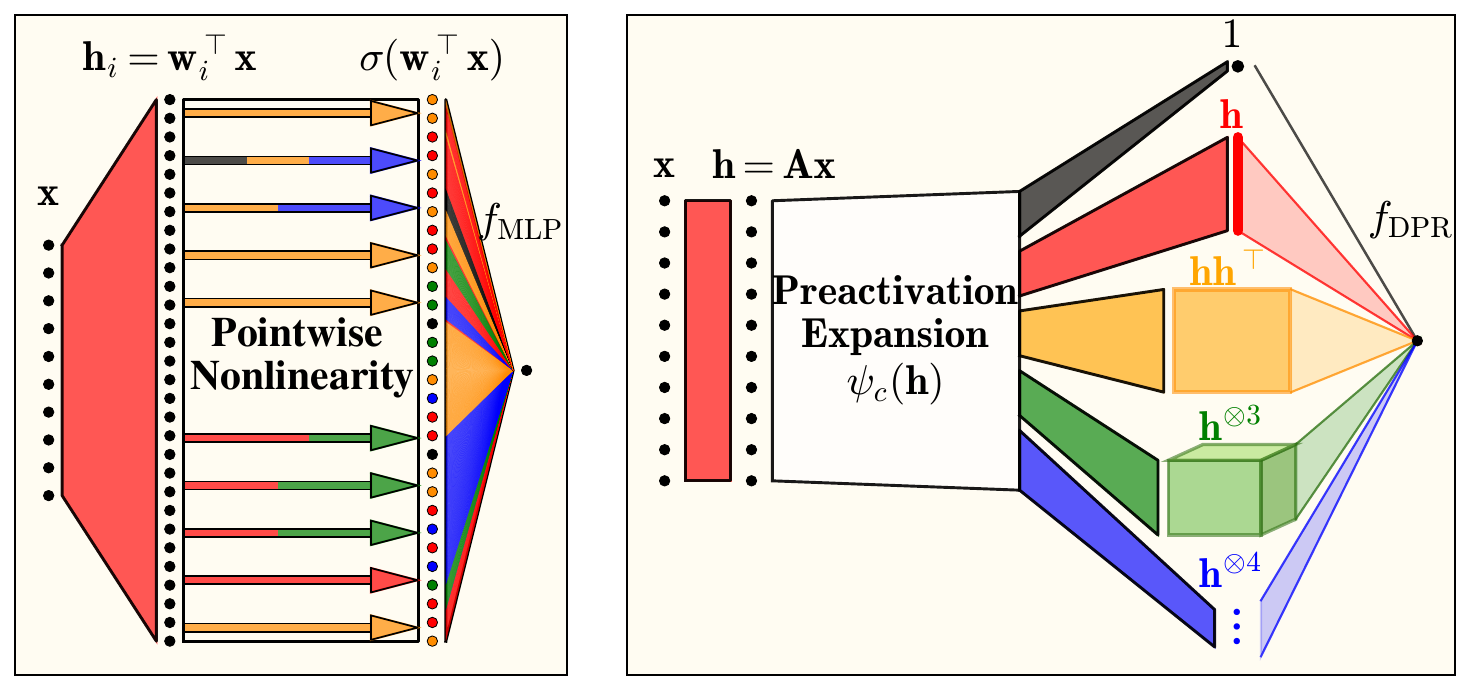}
    \caption{
    \textbf{Comparison between a shallow MLP and deep polynomial regression (DPR).} \textbf{Left.} A shallow MLP has nonlinearities applied elementwise, and each nonlinearity contains many polynomial orders. Different elements tend to different components of the preactivation. \textbf{Right.} In DPR, the nonlinear operation is a preactivation expansion which separately generates monomial interactions of the input vector.
    }
    \label{fig:model_description}
\end{figure}

Why is this a reasonable proxy for a shallow MLP?
The matrix $\mA$ serves the role of $\mW_\text{in}$, preconditioning the input to the nonlinear operation.
The operation $\vpsi_\vc$ generates all monomial functions of the hidden vector $\vh = \mA \vx$, giving the model full expressivity.
Finally, the readout vector $\mB$ maps these features to a scalar output.
See \cref{fig:model_description} for a visual comparison between the two models.

Note that if one simply fixes $\mA = \mI$, then \cref{eqn:dpr_functional_form} is polynomial regression: $f(\vx) = \langle \vb, \vpsi_\vc(\vx) \rangle.$
Equivalently, it is a kernel machine with kernel $K(\vx, \vx') = \sum_\ell \frac{c_\ell}{\ell!} \langle \vx, \vx' \rangle^\ell$.
With $\mA$ present, $f_\text{DPR}$ is a kernel machine with \textit{trainable} kernel $K_\mA(\vx, \vx') = \sum_\ell \frac{c_\ell}{\ell!} \langle \mA \vx, \mA \vx' \rangle$. 

The wonderful thing about DPR is that its gradient flow dynamics on squared loss can in many cases be reduced to a simple ODE of scalar variables and solved exactly.
The punchline of this paper is that the resulting ODE is often predictive of the loss trajectories and learning times for MLPs learning low-order polynomials (\cref{fig:main_figure,fig:staircase_losses}).

\section{Related works}\label{app:related-works}

Our work intersects with a wide range of preexisting literature, most notably: the analysis of single- and multi-index target functions, saddle-to-saddle dynamics, staircase function learning, mean-field treatments of two layer networks, and kernel learning.
We also briefly discuss potential relations to the ``neural feature matrix'' line of study.

\paragraph{Single- and multi- index models}

Single- and multi- index models (SIMs, MIMs respectively) have been proposed as the simplest non-trivial targets for shallow MLPs to learn.
Therefore, a great deal of literature covers how MLPs are trained on them \citep{benarous2021online,bietti:2022-single-index-models,mousavi2023gradient,damian2023smoothing,damian2022neural}.
Similar to our experiments on staircase functions, \citet{berthier2023timescales} studied the gradient flow trajectory of wide networks in the mean-field regime and found the same multi-timescale plateau-and-drop structure we showcase. 
Similar to our ODE, \citet{arnaboldi2023unifying, arnaboldi2024online} derived a dimensionless ODE detailing the batch size an MLP requires to escape a saddle point when trained on a SIM.
We occasionally induce a separation of timescales argument to reduce our feature ODE, which is preceeded by \citet{dandi2023onegiant,damian2023smoothing,dandi2024repetita,mousavi2025langevin} making batch size arguments for when this is valid on MIMs.

\paragraph{Saddle-to-saddle and step-like learning dynamics}

The empirical observation that gradient flow on small-initialization MLPs proceeds via long plateaus punctuated by rapid drops has by now been documented from many angles: saddle-to-saddle dynamics in deep linear networks \citep{jacot2021saddle}; plateaus and timescale separation in two-layer SIMs \citep{berthier2023timescales}; the implicit bias perspective of \citet{pesme2021implicit}; alternating gradient flows \citep{kunin2025alternatinggradientflowstheory}; and the structured plateaus exhibited by the teacher-student ODEs of \citet{saad1995online}.
Our feature ODE reproduces this plateau structure (\cref{fig:staircase_losses}) without explicitly solving for saddles.

\paragraph{Staircase functions}

The staircase functions studied in our experiments ($f^\star = z_1$, $z_1 + z_1z_2$, $z_1 + z_1z_2 + z_1z_2z_3$) come from the line initiated by \citet{abbe2021staircase,abbe2022merged,abbe2023sgdlearningneuralnetworks}.
Our work differs by analyzing and characterizing the effects of the anisotropy on the learning dynamics.
Closely related, the global convergence results of \citet{boursier2022relu} for ReLU networks on orthogonal inputs also exhibit a staircase decomposition of the loss.

\paragraph{Mean-field analyses of two-layer networks}

Our DPR model and feature ODE can be compared against the more rigorous yet less intuitive mean-field literature of \citep{mei2018mean,chizat2018global,sirignano2020mean,rotskoff2018interacting}.
At a high level, mean-field studies derive a \emph{partial} differential equation for the distribution of parameters in the $N\to\infty$ limit and gives global-convergence results under suitable assumptions.
These analyses take a slighly different approach to removing the effect of neuron specialization, taking an infinite-dimensional measure over neurons.
Close historical analogs to our ODE can be found from \citet{saad1995online,saad1995exact,goldt2019dynamics}, deriving order-parameter ODEs for teacher-student target functions (similar to our explicit Hermite target decomposition).
\citet{refinetti2021classifying} used analogous ODE reductions for high-dimensional classification problems.
Perhaps most similar to our work is the DMFT literature of \citet{bordelon2022self, bordelon2024finite} that captures the full evolution of inner-product kernels during feature learning.


\paragraph{Effective ODEs and high-dimensional limit theorems for SGD}

A separate analytical thread, instead of taking a width limit, takes a high-dimensional input limit and derives rigorous ODE/SDE descriptions of summary statistics under one-pass SGD.
\citet{collins2023hitting} established that online gradient flow on MIMs with general data covariance, admits an ODE that exactly predicts the risk and other observables.
\citet{arnaboldi2023unifying} gave a parallel result for two-layer networks, providing a dimensionless ODE that unifies mean-field and high-dimensional descriptions.

\paragraph{Kernels, Hermites, and random features}

Several of the technical ingredients we use---the Hermite decomposition of activations to read off the level coefficients $c_\ell$ (\cref{app:level_coeffs}), the dot-product kernel approximation in the ``near-origin'' or ``on-sphere'' regime, and the polynomial truncation of the random-feature kernel are similar to results found in \citep{karkada2025predictingkernelregressionlearning,wortsman:2025-krr-spectrum-with-powerlaw-data}.
The semi-frozen network discussed in \cref{app:mlp_to_expandernet} can be viewed as a random feature approximation to the DPR model; we owe \citet{rahimi:2007} for their random feature work.
\citet{mei2022generalization} used Hermite expansions to characterize precisely when neural networks beat their associated kernels, providing the polynomial-by-polynomial picture of approximation that our DPR truncation order $p$ inherits.

\paragraph{Neural feature matrix and feature-learning mechanisms}

We believe DPR is a sound tool to accurately describe the evolution of the neural feature matrix (NFM) $\mW_{\mathrm{in}}^\top \mW_{\mathrm{in}}$, introduced in \citep{radhakrishnan:2024-rfms}.

\paragraph{Preconditioned kernel learning}

We are not the first to suggest applying a learned preconditioner to the inputs of a kernel machine.
\citet{wilson2015deepkernellearning} first proposed this technique, using it to improve the sample complexity of kernel regression.

\section{The feature ODE}
\label{sec:ode}

We draw data as $\vx \sim \N(\mathbf{0}, \mGamma)$, with anisotropic covariance $\mGamma = \text{diag}(\gamma_1, \cdots, \gamma_d)$.
Given a target function $f^*$, we construct the square loss
$\L = \frac{1}{2} \E{\vx \sim \N(0,\mGamma)}{\left(f^*(\vx) - f_\text{DPR}(\vx) \right)^2}$.
We study gradient flow optimization of the DPR parameters:
\begin{equation}
    \dot{\mA} = - \nabla_\mA \L,
    \qquad
    \dot{\mB} = - \nabla_\mB \L,
    \qquad
    \mA(0) = 1 / \sqrt{d} \cdot \mI,
    \qquad
    \mB(0) = \mathbf{0}.
\end{equation}
Observe that the model function $f_\text{DPR}$ can be written out explicitly in polynomial form:
\begin{equation}
    f_\text{DPR}(\vx) =
    \sum_{\ell \ge 0}
    \sum_{\vbeta \in [d]^\ell}
    \frac{\sqrt{c_\ell}}{\sqrt{\ell!}}
    B_\vbeta
    \prod_{m = 1}^\ell
    (\mA \vx)_{\beta_m}.
\end{equation}
The great simplifying blessing is that, when $f^*$ depends on only a handful of the coordinates of $\vx$, then most of these terms can be neglected.
In particular, if $f^*$ depends on only a subset of coordinates $\{x_i\}$ such that $i \in \mathcal{I} \subset [d]$, then any readout parameters $b_\vbeta$ with $\vbeta \notin \mathcal{I}^d$ will remain identically zero under gradient flow.
Furthermore, first-layer parameters $A_{ij}$ with $j \notin \mathcal{I}$ will feel no gradient and remain equal to $\kappa \delta_{ij}$.

We now invoke two assumptions on the data.
First, we assume that $\gamma_i \ll 1$.%
\footnote{In our typical scaling, $\text{Tr}[\mGamma] = 1$ is the natural normalization, so this amounts to an assumption that no single coordinate gets a substantial fraction of the variance.}
With this assumption, we find that terms of higher orders are learned more slowly, and only the lowest-order terms of $f_\text{DPR}$ required to represent the target function $f^*$ grow substantially and are relevant to the dynamics.
Second, we expand the target function as a sum of Hermite polynomials: $f^*(\vx) = \sum_{\valpha \in \mathbb{N}_0^d} v_\valpha h_\valpha(\mGamma^{-1/2} \vx)$, and make a sparseness assumption on the coefficients $v_\valpha$ which will be satisfied in every case we discuss (the per-order parity criterion, \cref{app:parity}).
See \cref{app:approx_chain} for full discussion of these assumptions and the consequent derivations.
 
%
After some algebra, we are left with the following ODE:


\begin{align}
    \mathcal{L} &= \frac{1}{2}
    \sum_{\valpha \in \mathcal{N}_0^d}
    \left(
    v_\valpha -
    \tilde{c}_{\valpha}
    \cdot
    b_\valpha
    \prod_i a_i^{\alpha_i}
    \right)^2,
    \\
    \dot{a}_i &= -\partial_{a_i} \mathcal{L}, \qquad \qquad \dot{b}_{\valpha} = -\partial_{b_{\valpha}} \mathcal{L}\;,
\end{align}
where $\tilde{c}_\valpha := \sqrt{c_{|\valpha|}} \cdot \prod_{i=1}^d \gamma_i^{\alpha_i/2}$.
We refer to the above as the \textit{feature ODE}.
The point of the feature ODE is that (a) when the target function is very sparse (i.e. only a few $v_\alpha$ are nonzero), then we need only track a handful of scalar variables, and (b) the resulting loss dynamics resemble those of a shallow MLP trained on the same target function.

When the target function consists of a single Hermite term $f^*(\vx) = h_\valpha(\vx)$, the loss in the feature ODE simplifies to just $\L = \frac{1}{2} (1 - \tilde{c}_\valpha \cdot b \prod_i a_i^{\alpha_i})^2$, and the resulting dynamics of a few variables can be studied nicely.
The loss and its dynamics cannot be directly solved, but a number of approximations can be made, and are done so in \cref{app:clo}, such that an explicit solution for the effective optimization time $\tau \equiv \eta\cdot n_{\text{iter}}$ can be found.
%
In short, we define the effective optimization time by having $b$ reach its final value, of $b_f = \tilde{c}_{|\valpha|}^{-\frac{1}{\sum_i \valpha_i+1}}$:
\begin{equation}
    \int_0^\tau \text{d}t = \tau = \int_0^{b_{f}}\frac{\text{d}b}{\tilde{c}_{|\valpha|}\prod_{i=1}^d\left(\valpha_ib^2(t)+a_i^2(0)\right)^{\frac{\valpha_i}{2}}}\,.
\label{eq:rise_time}
\end{equation}
We experimentally validate that MLPs learn target functions in an amount of gradient steps defined by \cref{eq:rise_time}, finding great agreement with results summarized in the right side of \cref{fig:main_figure}.

\begin{figure}[t]
    \centering
    \includegraphics[width=\textwidth]{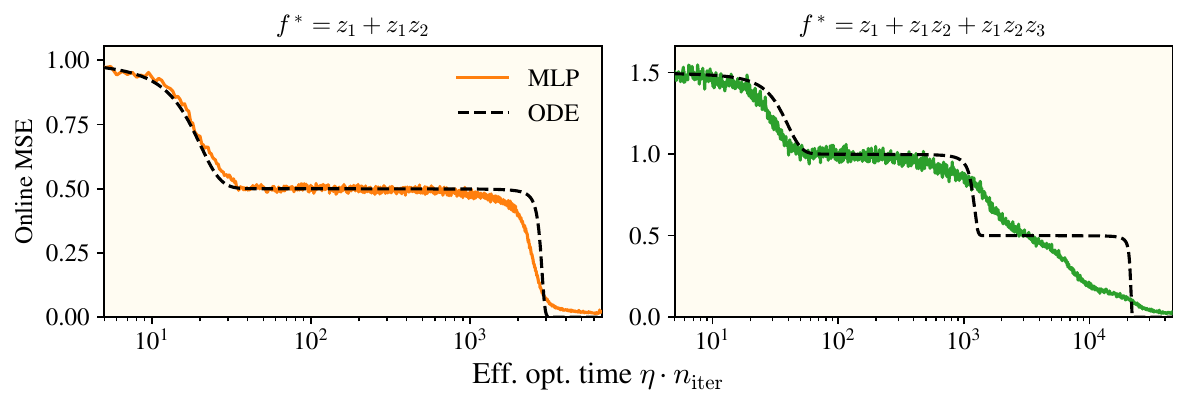}
    \caption{\textbf{The feature ODE predicts loss trajectories for shallow MLPs learning low-degree ``staircase functions.''} Predicted loss trajectories are staircases with a number of steps equal to the order of the target function. True loss curves for shallow MLPs roughly follow these staircases, with steps at roughly the predicted times. See \cref{app:experimental_setup} for experimental details.}
    \label{fig:staircase_losses}
\end{figure}

\paragraph{Example calculations.}

To concretize our findings, we provide an in-depth discussion on how the feature ODE can be used to analyze an MLP learning a number of targets presented in the paper.
Throughout our discussion, we will use normalized coordinates $z_i = x_i/\sqrt\gamma_i$ such that $|f^\star| = 1$.

\paragraph{Linear target.}

On a linear target, the feature ODE and dynamics reduce to the form
\begin{equation*}
    \mathcal{L} = \frac12\left( 1 - \tilde{c}_1 a_i b \right)^2 \qquad
    \dot{a}_i = (1-\tilde{c}_1a_ib)\cdot \tilde{c}_1b  \qquad \dot{b} = (1-\tilde{c}_1a_ib)\cdot \tilde{c}_1a_i\,.
\end{equation*}
where we would like to pause to appreciate the simplicity of our ODE:
describing the learning dynamics can reliably be captured by only \textit{two} dynamical variables!
For the remainder of the examples, we leave out the dynamics in $a$ and $b$ as they follow from simple derivatives of the loss.
For the loss trajectory, refer to the top left of \cref{fig:main_figure}, with the effective optimization time being
\begin{equation*}
    \tau = \int_0^{\tilde{c}^{\frac12}}\frac{\text{d}b}{\tilde{c}_1 \sqrt{b^2(t)+a_i^2(0)}} \approx \frac{1}{\tilde{c}_1}\ln\left(\frac{2}{\sqrt{\tilde{c}_1}a_i(0)}\right)
\end{equation*}
\begin{figure}[t]
    \centering
    \includegraphics[width=\textwidth]{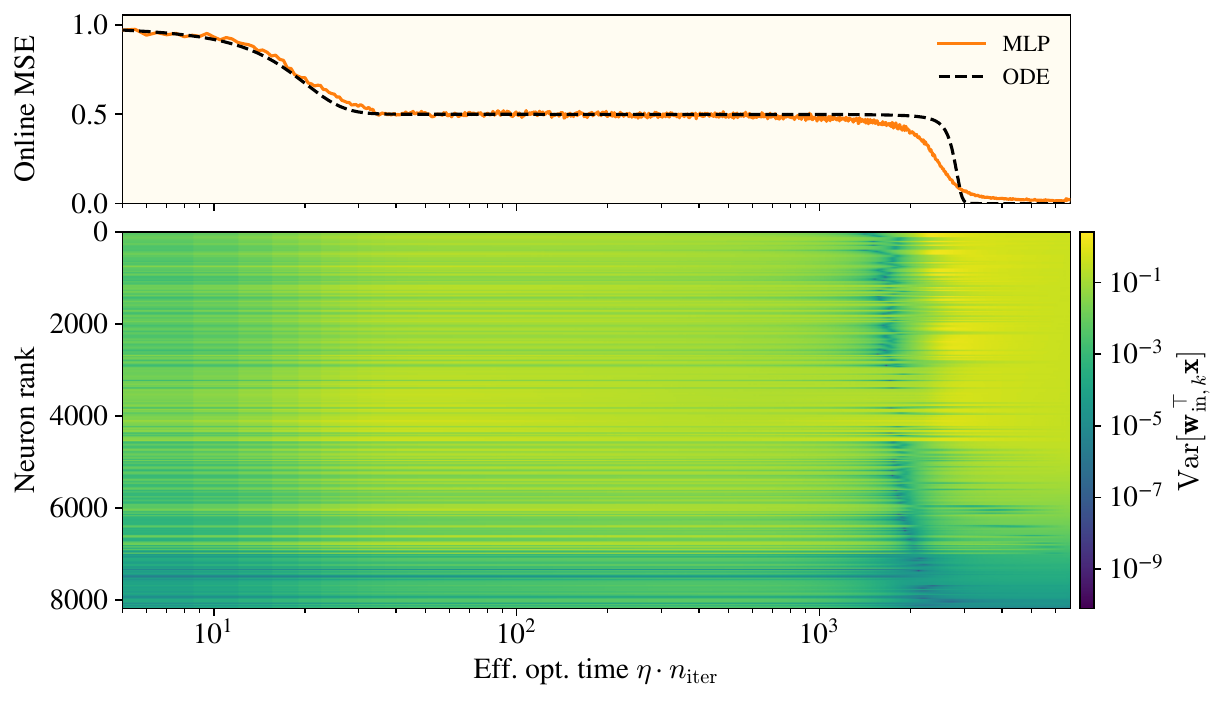}
    \caption{\textbf{The feature ODE can predict MLP loss dynamics despite neuron-level specialization in MLPs.} We train a two-layer MLP on a two-term staircase $f^*(\vz) = z_1 + z_1 z_2$ and overlay the feature ODE prediction.
    \textbf{Top.} We plot the online loss of the MLP and of the feature ODE.
    \textbf{Bottom.} Each row of the heatmap tracks one neuron's preactivation variance over training, and neurons are sorted by their variance at the end of training.
    The horizontal striations show neuron-level specialization: individual neurons carry distinctly different preactivation variances.
    Nevertheless, the feature ODE captures the MLP's staircase loss.}
    \label{fig:neuronspec}
\end{figure}

\paragraph{Quadratic targets.}

There are two different types of quadratic targets to consider: 
the first consists of a mixed-index target, $z_iz_j$, and the second is a pure-quadratic target $h_2(z_i)$.
We review the pure-quadratic case's effective optimization time, with the mixed-index being developed similarly.
We note that the feature ODE can be obtained by similar arguments used for the linear case,\footnote{Note that, strictly speaking, the pure-quadratic ODE lacks a corresponding mode to the constant $1$. 
So long as $\tilde{c}_2 \ll \tilde{c}_0$, the difference in dynamics when this term is excluded compared to included is negligible.}
We find the dynamics and effective optimization time as,
\begin{align*}
    \tau &\approx \int_0^{\infty}\frac{\text{d}b}{\tilde{c}_2\left(2b^2(t)+a_i^2(0)\right)} = \frac{a_0^{-1}}{\tilde{c}_2} \frac{\pi}{2^{3/2}}\,,
\end{align*}
The mixed-index case has one small but noticeable change: the effective optimization time has the $2^{3/2}$ in the denominator replaced with a $2$.
Surprisingly, the repeated-index quadratic is learned with exactly $\sqrt{2}$ times fewer steps than it would take to learn a mixed-index quadratic, all other properties remaining constant (validated in \cref{fig:main_figure})!

\paragraph{Staircase with two terms}

Thus far, we have only discussed target functions with a single term.
Notably, this is vastly different from what most real targets consist of.
To get closer to real targets, we analyze feature ODE on a ``staircase function'' (defined in \citet{abbe2023sgdlearningneuralnetworks}) with two terms: $f^\star = z_i+z_iz_j$.
The dynamics of the feature ODE are not discussed in detail, however they follow similar arguments to those of the linear and quadratic cases and are instead shown in \cref{fig:staircase_losses}.
We note that there is currently no clean expression for the optimization time of the feature ODE on staircase functions, although presume it is possible.


As more terms are added, the feature ODE will undoubtedly become more complex, however we believe the feature ODE will still be a useful tool that can be used to analyze the learning dynamics of the shallow MLP.
\section{The feature ODE cannot describe neuron-level specialization}
\label{sec:neuron}

Although our models are fully expressive and capture the training dynamics of simple target functions well, we cannot fully capture all of the training dynamics of MLPs.
We removed the elementwise nonlinearity and replaced it with a global nonlinear operation.
With each neuron being indistinguishable, we have thus eliminated all neuron-level specialization.
MLPs trained on general tasks, however, are known to develop sparse, specialized weights (i.e.\ some rows $\vw_{\text{in},k}$ of $\mW_\text{in}$ satisfy $\|\vw_{\text{in},k}\|_1 \ll \sqrt{d}\,\|\vw_{\text{in},k}\|_2$).
Select neurons can attend to few input coordinates or directions \citep{maennel2018gradient, barak2022hidden, boursier2024early}, notably when developing localized receptive fields on image data \citep{ingrosso2022data, lufkin2024nonlinear} or solving algorithmic tasks like modular addition \citep{gromov2023grokking, nanda2023progressmeasuresgrokkingmechanistic, kunin2025alternatinggradientflowstheory, he2026mechanism}.
Our model is unable to capture these neuron-level dynamics.
For simple target functions such as the staircase in \cref{fig:neuronspec}, the effects of neuron-level specialization are minor and the feature ODE can predict the MLP's loss well even though MLP neurons clearly specialize.
For other target functions, specialization becomes significant, and the feature ODE predicts the loss less accurately.
See \cref{app:neuron_spec} for details.

\section{Outlook \& Conclusion}
\label{sec:conclusion}

We are excited about future work utilizing this framework to study feature learning.
We believe our framework allows for a principled study of the evolution of the neural feature matrix $\mW_{\text{in}}^\top\mW_{\text{in}}$ (introduced in \citet{radhakrishnan:2024-rfms}), with some, albeit weaker, evidence for a principled study of the output layer weight vector $\vw_{\text{out}}$ being possible.
We are optimistic we can predict dynamics of MLPs learning more complicated functions in varied hyperparameter regimes by further studying DPR and studying more general simplified ODEs.

\section*{Acknowledgements}
\label{sec:acknowledgements}

All authors were supported by the FLAB initiative at Imbue and are grateful for Imbue's support of fundamental science.
All authors are grateful to Dhruva Karkada for extensive discussions about the project.

JT thanks the 2025 learning mechanics retreat group for the realization of the promising potential for the feature ODE, along with the space given to meet with co-authors in-person.
BO acknowledges support by the National Science Foundation Graduate Research Fellowship Program.
BO is also grateful to Surbhi Goel for extensive discussion about the project.
%

\bibliography{refs}

\clearpage
\appendix

\section{Further experimental details}
\label{app:experimental_setup}

For the MLP, we use $\mu$P parameterization \citet{yang:2021-tensor-programs-IV}:
$$
\mW_{\text{in}}[n,i] \sim \N(0, 1/d), \quad \mW_{\text{out}}[n] = \textbf{0}, \quad \eta_{\mW_\text{in}} = \eta\, N, \quad \eta_{\mW_\text{out}} = \eta/N,
$$
where $\mW_{\text{in}} = \mathbb{R}^{N \times d}$, and we set $\mW_{\text{out}}$ to be the all zero's vector for consistency with the feature ODE (as $N\to \infty$, $\mW_{\text{out}}$ tends to 0).
For the SFN and DPR, we initialize $\mA = \frac{1}{\sqrt{d}}\mI_d$ and zero out the readout tensors $\mB^{(\ell)}$.
All three models train with online SGD on streamed Gaussian data.

\paragraph{\cref{fig:main_figure} experimental details.}

Experiments are performed with data $\vx \sim \mathcal{N}(0, \Gamma)$, with $\vx \in \mathbb{R}^{100}$ and $\Gamma = \text{diag}(\gamma_i)$ where $\gamma_i \sim (i+6)^{-1.8}$ normalized such that $\sum_i\gamma_i = 1$.
The learning rate is set to $\eta = 10^{-2}$ which we find to be near the gradient flow regime.
Our data is streamed online with a batch size $bsz=4000$---we note that any reasonably large choice for the batch size reliably reproduces our results.
Our MLP is setup in the $\mu$P regime \citet{yang:2021-tensor-programs-IV}, with a width $N=8192$.
Our activation function of choice is a shifted SiLU: $\sigma(x) = \text{SiLU}(x+0.5)$ which offers the nice properties of (1) being relatively close to the standard practice ReLU, and (2) having a full power series, with the Taylor expansion around $x=0$ having no coefficients of 0.
We make explicit note that the vast majority of biasless nonlinearities are typically not full in their power series.
An exponential moving average smoother with constant $0.9$ ($\mathcal{L}(t) = 0.9\mathcal{L}(t-1)+0.1\Delta\mathcal{L}(t)$ is used to smooth over random Gaussian fluctuations in the data.
\textbf{Loss curves.}
Training is done with a set $100,000$ gradient steps, at which point the run terminates.
\textbf{Effective optimization time predictions.}
We train on a number of targets finding great agreement on all linear terms with slight disagreement on quadratic terms with high covariance $\gamma_i$.
Cubic and higher order target functions are not included as the predicted optimization times were prohibitive.
We believe anisotropic effects that make real MLP optimization times differ than the ODE predicted times are more prevalent for these higher order targets.

\paragraph{\cref{fig:staircase_losses} experimental details.}

Both panels train an MLP of width $N = 8192$ online with batch size $10{,}000$ and a single init seed, and overlay the feature ODE prediction.
Panel~(a) targets the 2-staircase $f^*(\vz) = z_1 + z_1 z_2$ with $d = 2$, a power-law exponent $\alpha = 10$ yielding $\bm{\gamma} \approx (2.00\times 10^{-2},\, 1.95\times 10^{-5})$, learning rate $\eta = 0.5$, and cosine activation $\sigma(z) = \sqrt{2}\cos(z + \pi/4)$.
Panel~(b) targets the 3-staircase $f^*(\vz) = z_1 + z_1 z_2 + z_1 z_2 z_3$ with $d = 3$, a power-law exponent $\alpha = 6$ yielding $\bm{\gamma} \approx (2.95\times 10^{-2},\, 4.61\times 10^{-4},\, 4.05\times 10^{-5})$, learning rate $\eta = 0.03$, and tanh-half activation. We were only able to see the staircase behavior with a large ratio of data eigenvalues. With larger data eigenvalues and smaller learning rate the feature ODE becomes less predictive.

\paragraph{\cref{fig:val_sfn} and \cref{fig:val_dpr} experimental details.}

Both figures share the same configuration: $d = 3$, a power-law exponent $\alpha = 1.5$ yielding $\bm{\gamma} \approx (3.11\times 10^{-3},\, 1.10\times 10^{-3},\, 5.98\times 10^{-4})$, the cosine activation $\sigma(z) = \sqrt{2}\cos(z + \pi/4)$ with no biases, learning rate $\eta = 0.2$, batch size $10{,}000$, and a single init seed.
The SFN uses width $N = 8192$, and DPR uses truncation order $p = 3$ to accommodate the highest-order target.
We sweep across four targets in $\vz$ coordinates: $z_1$, $\mathrm{He}_2(z_1)$, $z_1 + z_1 z_2$, and $z_1 + z_1 z_2 + z_1 z_2 z_3$.

\paragraph{\cref{fig:dyn_linear}, \cref{fig:dyn_xnorm}, and \cref{fig:dyn_staircase2} experimental details.}

Each figure compares the diagonal-$\mA$ dynamics of an MLP run against an SFN run on the same target.
All six runs use width $N = 8192$, batch size $10{,}000$, the cosine activation $\sigma(z) = \sqrt{2}\cos(z + \pi/4)$ with no biases, and a single init seed.
The per-run hyperparameters are:
\begin{itemize}
    \item \cref{fig:dyn_linear}, target $f^*(\vz) = z_1$: MLP uses $d = 2$ with uniform $\bm{\gamma} = (2.50\times 10^{-3},\, 2.50\times 10^{-3})$ and $\eta = 0.1$; SFN uses $d = 3$ with $\alpha = 1.5$ yielding $\bm{\gamma} \approx (3.11\times 10^{-3},\, 1.10\times 10^{-3},\, 5.98\times 10^{-4})$ and $\eta = 0.2$.
    \item \cref{fig:dyn_xnorm}, target $f^*(\vz) = \mathrm{He}_2(z_1) + \mathrm{He}_2(z_2)$: MLP and SFN both use $d = 2$ with uniform $\bm{\gamma} = (2.50\times 10^{-3},\, 2.50\times 10^{-3})$ and $\eta = 0.1$.
    \item \cref{fig:dyn_staircase2}, target $f^*(\vz) = z_1 + z_1 z_2$: MLP uses $d = 2$ with $\alpha = 10$ yielding $\bm{\gamma} \approx (2.00\times 10^{-2},\, 1.95\times 10^{-5})$ and $\eta = 0.5$; SFN uses $d = 3$ with $\alpha = 1.5$ yielding $\bm{\gamma} \approx (3.11\times 10^{-3},\, 1.10\times 10^{-3},\, 5.98\times 10^{-4})$ and $\eta = 0.2$.
\end{itemize}

\section{Detailed derivation of the feature ODE from MLPs}
\label{app:approx_chain}

We derive the feature ODE from an MLP with MSE loss in three steps.
Rather than conjecturing DPR directly from the MLP, we introduce an intermediate model---the \emph{semi-frozen network} (SFN)---whose infinite-width limit recovers DPR, and we derive the SFN from the MLP by a conjecture on the input layer's weight matrix.

\subsection{Further details on MLPs to Semi-Frozen Networks}
\label{app:mlp_to_expandernet}

A one-hidden-layer MLP computes
$$
f_{\text{MLP}}(\vx) = \mW_{\text{out}}\, \sigma(\mW_1 \vx), \qquad \mW_1 \in \R^{N \times d},
$$
where $\sigma$ acts pointwise.
To pass from the MLP to a semi-frozen network, we assume that the hidden neurons form a homogeneous population: no single neuron encodes special information, and adding or removing any one has a negligible effect on the output.
We require this homogeneity to hold throughout training, not just at initialization.

The input layer plays two roles.
It rotates the data, upweighting coordinates that matter for the output; and it routes the rotated data into the pointwise nonlinearity, where different neurons see different mixtures.
The SFN disentangles these roles by freezing the routing matrix and training only the rotation.

\paragraph{Skinny SVD.}
Let $\mW_1 = \mU\mS\mV^\top$ be the thin SVD, with $\mU \in \R^{N \times d}$ having orthonormal columns, $\mS \in \R^{d \times d}$ diagonal, and $\mV \in \R^{d \times d}$ orthogonal.
Defining $\mA := \mS\mV^\top$ and $\mW_{\text{frozen}} := \mU$ gives
$$
f_{\text{SFN}}(\vx) = \mW_{\text{out}}\, \tilde{\sigma}(\mW_{\text{frozen}} \mA \vx),
$$
where $\tilde{\sigma}(x)=\sum_{\ell} q_\ell c_\ell x^\ell$, $c_\ell$'s given by the Taylor series expansion of the MLP's $\sigma$ (see \cref{app:level_coeffs}).
We introduce a factor of $q_\ell$ as a numerical parameter which we found to follow the pattern of $0: 1, 1:1, 2:2$ for our experiments.
We suspect $q_\ell = \ell!$, although we provide no such proof.
$\tilde{\sigma}$ still acts pointwise, but now on the random features $\mW_{\text{frozen}} \mA \vx$.
We initialize $\mA = \frac{1}{\sqrt{d}}\mI_d$ to remain consistent with $\mu$P scaling.

The SFN \emph{freezes} $\mU$ at its initialization and trains only $\mA$, reducing the first-layer parameter count from $Nd$ to $d^2$.
In practice, rather than taking $\mU$ from the SVD of a specific $\mW_1$, we initialize each entry of $\mW_{\text{frozen}}$ as iid Gaussian with variance $\sigma_{\text{frozen}}^2/d$ (the standard $\mu$P / random-feature scaling) and hold it fixed throughout training.
The scale $\sigma_{\text{frozen}}$ is a free hyperparameter that controls the preactivation magnitude; we absorb it into the data eigenvalues $\gamma_i$ throughout (see the footnote in \cref{app:expandernet_to_dpr}), so it does not appear explicitly in the kernel or the feature ODE.
This shift assigns the nonlinearity's expressive role to a fixed random projection.
The price is any reorganization of the frozen directions---basis rotations, ICA-style updates---which we discuss in \cref{app:neuron_spec}.

\paragraph{Width and concentration.}
As $N \to \infty$, the SFN's random feature kernel concentrates around its expectation over the frozen weights, which is the SFN kernel we use in \cref{app:expandernet_to_dpr}.
At finite $N$, the SFN's Gram matrix fluctuates by $O(1/\sqrt{N})$ across seeds, so the SFN matches the kernel only up to width corrections.

\paragraph{Neural feature matrix.}
The MLP carries a ``neural feature matrix'' $\text{NFM} := \mW_{\text{in}}^\top \mW_{\text{in}}$.
For an SFN, $\mA$ plays the analogous role:
$$
\mA \;\sim\; \sqrt{\tfrac{1}{N}\, \mW_{\text{in}}^\top \mW_{\text{in}}}.
$$

\subsection{Further details on Semi-Frozen Networks to DPR}
\label{app:expandernet_to_dpr}

Here, we replace the pointwise activation in the SFN with explicit polynomial features and obtain deep polynomial regression (DPR).
At a high level, the SFN's feature kernel admits a polynomial decomposition, and DPR is the explicit-feature model that realizes the same decomposition.

\paragraph{The SFN feature kernel.}
Because $\mW_{\text{frozen}}$ is frozen, the random feature kernel holds throughout training, not only at initialization (unlike the NNGP).
Taking $N \to \infty$, the random feature average converges to its expectation over the frozen weights\footnote{Three equivalent choices set the preactivation scale and produce identical dynamics: (a) rescale the data eigenvalues $\gamma_i \to \beta^2 \gamma_i$, breaking the trace normalization $\sum_i \gamma_i = 1$; (b) shift to higher-indexed input coordinates of the same power-law spectrum, which yields smaller eigenvalues while preserving the trace; (c) multiply the preactivation by $\beta$ inside the model itself, equivalent to writing a $\sigma_{\text{frozen}}$ factor inside the activation. Throughout the appendix and the experimental details (\cref{app:experimental_setup}) we absorb any preactivation scale into the data eigenvalues $\gamma_i$, so the kernel below carries no $\sigma_{\text{frozen}}$ factor.} \citep{rahimi:2007}:
$$
K_{\text{SFN}}(\vx, \vx') = \E{\vw \sim \N(0, \mI_d / d)}{\tilde{\sigma}(\vw^\top \mA \vx)\, \sigma(\vw^\top \mA \vx')}.
$$
The $1/d$ per-component variance is the standard $\mu$P / random-feature scaling that keeps the preactivation $O(1)$ as $d$ grows.
The preactivations $\vw^\top \mA \vx$ and $\vw^\top \mA \vx'$ are jointly Gaussian, so the kernel reduces to a two-dimensional Gaussian expectation:
$$
K_{\text{SFN}}(\vx, \vx') = \E{z_1, z_2 \sim \N(0, \mSigma')}{\tilde{\sigma}(z_1)\, \tilde{\sigma}(z_2)}, \quad \mSigma' = \frac{1}{d} \begin{pmatrix} \norm{\mA \vx}^2 & \vx^\top \mA^\top \mA \vx' \\ {\vx'}^\top \mA^\top \mA \vx & \norm{\mA \vx'}^2 \end{pmatrix}.
$$
The kernel therefore depends on three scalars: the two norms $\norm{\mA\vx}^2$, $\norm{\mA\vx'}^2$ and the dot product $\vx^\top \mA^\top \mA \vx'$.

\paragraph{Dot-product approximation.}
The kernel depends on all three scalars in general, but reduces to a function of the dot product alone in two regimes.
First, when $\norm{\mA\vx}^2$ concentrates across data points (the ``on-sphere'' regime), the diagonal entries factor out as a data-independent prefactor.
Second, when $\norm{\mA\vx}^2/d$ stays small for all data points (the ``near-origin'' regime), the diagonal entries sit near zero and the kernel ignores their variation entirely.
The near-origin regime corresponds to the wide-kernel-theory limit; the on-sphere regime is weaker, allowing larger norms but requiring them to be approximately equal.
Small data eigenvalues place us in the near-origin regime, since $\E{\vx}{\norm{\mA\vx}^2/d} = (1/d)\sum_j a_j^2 \gamma_j$.
In both cases, the kernel reduces to
$$
K_{\text{SFN}}(\vx, \vx') \approx k_{\text{SFN}}\!\left(\tfrac{1}{d}\, \vx^\top \mA^\top \mA \vx'\right).
$$

\paragraph{Polynomial decomposition.}
Taylor-expanding $k_{\text{SFN}}$ gives
$$
K_{\text{SFN}}(\vx, \vx') \approx \sum_{\ell=0}^{\infty} \tfrac{c_\ell}{\ell!} \left(\tfrac{1}{d}\,\vx^\top \mA^\top \mA \vx'\right)^\ell \equiv K_{\text{DPR}}(\vx, \vx'),
$$
where the $(1/d)^\ell$ factors are absorbed into $c_\ell$.
Applying Mercer's theorem with eigenfunctions $\phi \sim \sqrt{c_\ell/\ell!}\, (\mA\vx)^{\otimes \ell}$ yields the DPR model:
$$
f_{\text{DPR}}(\vx) = \sum_{\ell=0}^{p} \sqrt{\tfrac{c_\ell}{\ell!}}\, \mB^{(\ell)} (\mA\vx)^{\otimes \ell},
$$
with trainable symmetric order-$\ell$ tensors $\mB^{(\ell)}$ and an order-$p$ truncation.

\paragraph{Truncation error.}
DPR drops all terms $\ell > p$ in the expansion above.
The order-$\ell$ kernel contribution is $\tfrac{c_\ell}{\ell!}\, (\tfrac{1}{d}\,\vx^\top \mA^\top \mA \vx')^\ell$, which decays with $\ell$ when $\norm{\mA\vx}^2/d$ stays small.
Taking the expectation over data $\vx \sim \N(0, \mGamma)$ gives $\E{\vx}{\norm{\mA\vx}^2/d} = (1/d)\sum_j a_j^2 \gamma_j$, so smaller data eigenvalues and larger $p$ both tighten the approximation.

\subsection{Further details on DPR to feature ODE}
\label{app:dpr_to_ode}

We simplify DPR's gradient flow ODE to a system of scalar variables in $\mA$ and the target-aligned entries of $\mB$, recovering the feature ODE defined in the main text.
The reduced state consists of $\mA$'s diagonal entries $\va = (a_1, \ldots, a_d)$ together with one scalar $b_\valpha$ per target Hermite mode---the common value of $\mB^{(|\valpha|)}$ on permutations of $\valpha$.

The reduction rests on the same two assumptions the main text invokes (\cref{sec:ode}): the data eigenvalues are small, which separates the timescales of different polynomial orders; and the target's nonzero Hermite coefficients $v_\valpha$ obey a per-order parity criterion, which keeps $\mA$ diagonal under gradient flow.

\paragraph{The reduced equations.}

Under both assumptions, only the target-aligned $\mB$ entries drive the dynamics, and we replace each permutation-symmetric block $\mB^{(|\valpha|)}$ on multi-index $\valpha$ by a single scalar $b_\valpha$.
Expanding $f_\text{DPR}$ around diagonal $\mA$ and computing the Hermite coefficient at each target mode $\valpha$ yields, after some algebra,
$$
\E{\vx}{f_{\text{DPR}}(\vx)\, h_\valpha(\mGamma^{-1/2}\vx)} = \tilde{c}_\valpha\, b_\valpha \prod_j a_j^{\alpha_j}, \qquad \tilde{c}_\valpha := \sqrt{c_{|\valpha|}} \prod_j \gamma_j^{\alpha_j/2},
$$
matching the main-text definition (\cref{sec:ode}).
The residual at mode $\valpha$ is $e_\valpha := v_\valpha - \tilde{c}_\valpha\, b_\valpha \prod_j a_j^{\alpha_j}$, and the population gradient flow produces the feature ODE:
$$
\L = \tfrac{1}{2}\sum_\valpha e_\valpha^2,
$$
$$
\dot{b}_\valpha = e_\valpha\, \tilde{c}_\valpha \prod_j a_j^{\alpha_j}, \qquad
\dot{a}_j = \sum_{\valpha:\, \alpha_j \geq 1} e_\valpha\, \tilde{c}_\valpha\, \alpha_j\, \frac{\prod_k a_k^{\alpha_k}}{a_j}\, b_\valpha.
$$
The sum runs over target multi-indices, since $v_\valpha = 0$ kills the residual elsewhere.

\paragraph{Parity criterion for diagonal $\mA$.}
\label{app:parity}

The off-diagonal $A_{ij}$ gradient takes the form $\E{\vx}{(f - f^*)\, \vx^{\valpha'}}$, with $\valpha'$ shifted by one in coordinates $i, j$.
Gaussian parity makes this expectation vanish unless every entry of $\valpha'$ is even, which in turn forces two target multi-indices at the same order to differ in parity at exactly $i, j$.
So $\mA$ stays diagonal whenever no two same-order target multi-indices differ in parity at exactly two coordinates---the \emph{per-order parity criterion}.
Staircases ($z_1$, $z_1 z_2$, $z_1 z_2 z_3, \ldots$) and single Hermite modes ($\mathrm{He}_\ell(z_1)$) satisfy this criterion, and we restrict to such targets throughout the paper.
Targets that fail include $z_1 + z_2$ and $\mathrm{He}_3(z_1) + \mathrm{He}_3(z_2)$, where two same-order multi-indices differ in exactly two coordinates and seed an off-diagonal gradient.

\paragraph{Timescale separation.}

The order-$\ell$ Hermite mode's gradient carries a factor of order $\gamma^{\valpha}$, so higher-order modes relax much more slowly than lower-order ones whenever the data eigenvalues are small.
Lower-order Hermite modes therefore settle long before higher-order modes have moved, and off-target $\mB$ entries don't backreact on the target dynamics at leading order in the eigenvalues.
We can therefore analyze each target multi-index independently---the property the reduced ODE above relies on.

\subsection{Validation of the approximation chain}
\label{app:approximation_validation}

\begin{figure}[H]
    \centering
    \includegraphics[width=\textwidth]{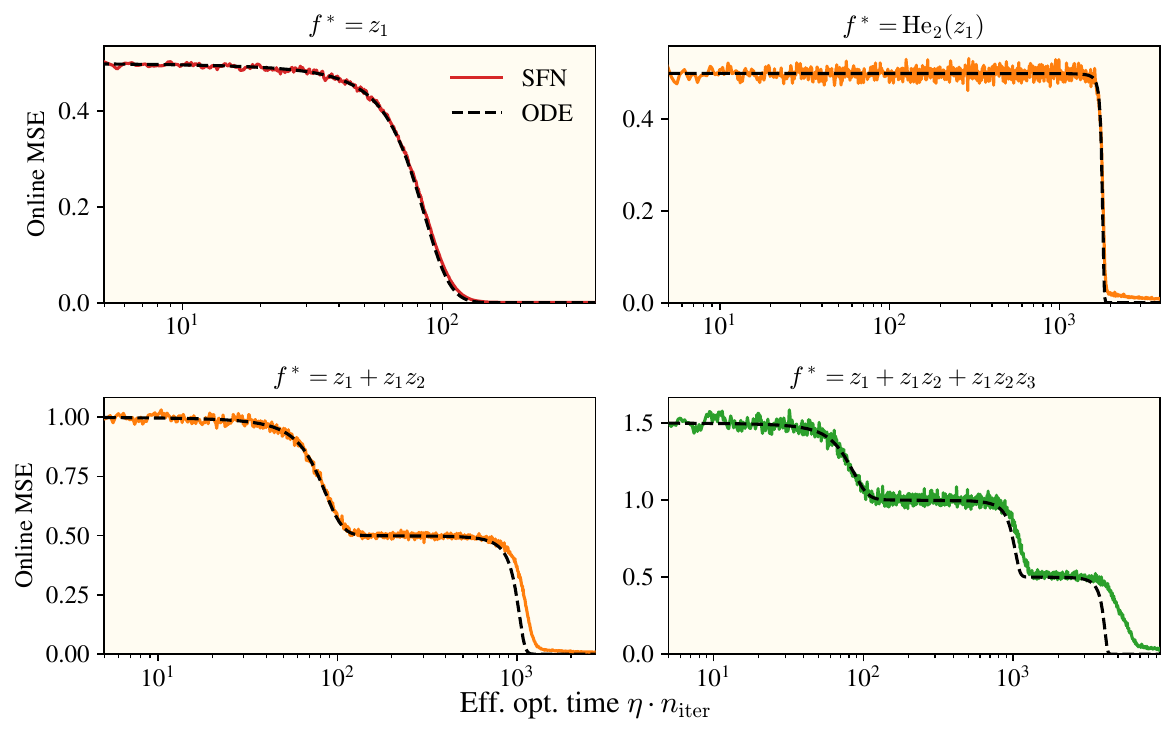}
    \caption{\textbf{SFN tracks the feature ODE across all four targets.} Online MSE for an SFN (colored) and the feature ODE (dashed black) over effective optimization time. The ODE recovers both the plateau lengths and the loss drops for each target, including the staircase structure of the two- and three-term staircase targets.}
    \label{fig:val_sfn}
\end{figure}

\begin{figure}[H]
    \centering
    \includegraphics[width=\textwidth]{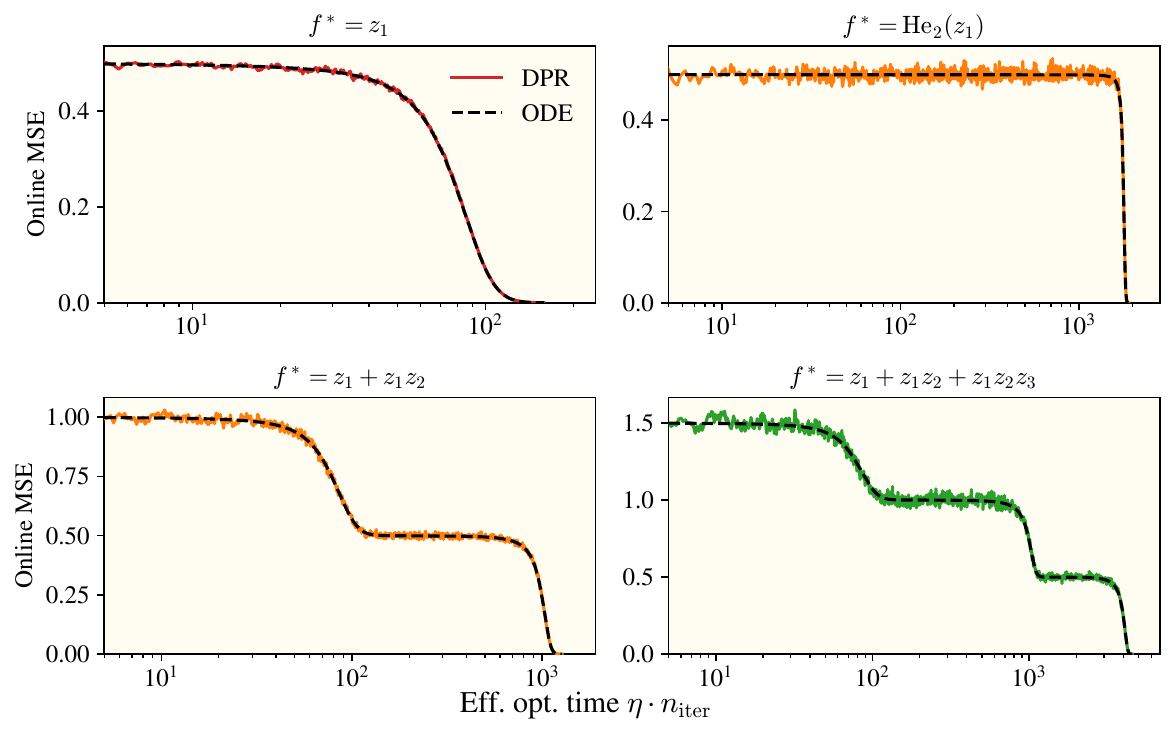}
    \caption{\textbf{DPR tracks the feature ODE across all four targets.} Same as \cref{fig:val_sfn} but with DPR replacing the SFN. The agreement is tighter than for the SFN, consistent with DPR's explicit polynomial features removing the random-feature concentration error.}
    \label{fig:val_dpr}
\end{figure}

We validate the two links of the approximation chain on synthetic Gaussian data: the SFN-to-feature-ODE reduction (\cref{fig:val_sfn}) and the DPR-to-feature-ODE reduction (\cref{fig:val_dpr}).
In both cases, the feature ODE is solved with the same hyperparameters as the trained network.
We compare online MSE across four targets of increasing complexity: a linear monomial $f^* = z_1$, a single Hermite mode $f^* = \mathrm{He}_2(z_1)$, a two-term staircase $f^* = z_1 + z_1 z_2$, and a three-term staircase $f^* = z_1 + z_1 z_2 + z_1 z_2 z_3$.
See \cref{app:experimental_setup} for the experimental setup.

\section{Level coefficients for general activations}
\label{app:level_coeffs}

\subsection{General formula}

For a general activation $\sigma$, with data on a sphere $\norm{\mA\vx} = r$, the level coefficients $c_\ell$ of the dot-product kernel can be computed from the Hermite decomposition of the rescaled activation:
\begin{equation}
    c_\ell = \frac{\ell!}{r^{2\ell}} \left(\E{z \sim \N(0,1)}{h_\ell(z)\, \sigma\!\left(\tfrac{r}{\sqrt{d}} \cdot z\right)}\right)^{\!2},
\end{equation}
where $h_\ell(z) = \frac{1}{\sqrt{\ell!}} \mathrm{He}_\ell(z)$ is the $\ell$-th normalized probabilist's Hermite polynomial. In words, we extract the $\ell$-th Hermite coefficient of the function $z \mapsto \sigma(r z/\sqrt{d})$, where the prefactor $r/\sqrt{d}$ is the standard deviation of the preactivations.

The activation's symmetry constrains which orders are present: odd activations (e.g.\ tanh) have $c_{2\ell} = 0$; activations that are linear plus even (e.g.\ ReLU) have $c_\ell = 0$ for $\ell = 3, 5, 7, \ldots$.
For the reason, we often use either a shifted activation function (e.g. SiLU$(x+0.5)$, $\tanh(x+0.5)$) or a cosine activation function ($\sigma(z) = \sqrt{2}\cos(z + \pi/4)$), all of which are full in their Taylor series.

\paragraph{Approximation.}

While we are in the near-origin regime (discussed in \cref{app:expandernet_to_dpr}), we find that the level coefficients are found from taking it's Taylor series expansion:

\begin{equation}
    c_\ell = \left(\sigma^{(\ell)}(0)\right)^2\,,
    \label{eq:level_coeff}
\end{equation}

the exponent $(\ell)$ in parentheses denoting the $\ell$-th derivative.
\cref{eq:level_coeff} gives the level coefficients used in our validation experiments.

\subsection{Cosine activation}

Here, we derive the dot-product kernel for the special case of a cosine activation $\sigma(z) = \sqrt{2}\cos(z + \pi/4)$ and show that it gives $c_\ell = 1$ for all $\ell$.

The SFN feature kernel is
$$
K(\vx, \vx') = \E{\vw \sim \N(0, \mI_d/d)}{\sigma(\vw^\top \mA \vx)\, \sigma(\vw^\top \mA \vx')}.
$$
The preactivations $z_1 = \vw^\top \mA \vx$ and $z_2 = \vw^\top \mA \vx'$ are jointly Gaussian with covariance
$$
\Sigma' = \frac{1}{d} \begin{pmatrix} \norm{\mA\vx}^2 & \vx^\top \mA^\top \mA \vx' \\ {\vx'}^\top \mA^\top \mA \vx & \norm{\mA\vx'}^2 \end{pmatrix}.
$$
Write $s_{11} = \frac{1}{d}\norm{\mA\vx}^2$, $s_{22} = \frac{1}{d}\norm{\mA\vx'}^2$, and $s_{12} = \frac{1}{d}\,\vx^\top \mA^\top \mA \vx'$.

Expanding the activation:
\begin{align}
K &= 2\,\E{}{\cos(z_1 + \pi/4)\cos(z_2 + \pi/4)} \\
  &= \E{}{\cos(z_1 - z_2)} + \E{}{\cos(z_1 + z_2 + \pi/2)}.
\end{align}
For the first term, $z_1 - z_2 \sim \N(0, s_{11} - 2s_{12} + s_{22})$. Using the Gaussian identity $\E{u \sim \N(0, \tau^2)}{\cos(u)} = e^{-\tau^2/2}$:
$$
\E{}{\cos(z_1 - z_2)} = \exp\!\left(-\tfrac{1}{2}(s_{11} - 2s_{12} + s_{22})\right).
$$
For the second term, $z_1 + z_2 \sim \N(0, s_{11} + 2s_{12} + s_{22})$, and $\cos(z_1 + z_2 + \pi/2) = -\sin(z_1 + z_2)$. Since $\E{u \sim \N(0,\tau^2)}{\sin(u)} = 0$ by symmetry, this term vanishes. Therefore:
$$
K(\vx, \vx') = \exp\!\left(-\tfrac{1}{2}(s_{11} - 2s_{12} + s_{22})\right) = \exp\!\left(-\tfrac{1}{2d}\norm{\mA\vx - \mA\vx'}^2\right).
$$
This is an RBF kernel in the transformed coordinates $\mA\vx$. Factoring:
$$
K(\vx, \vx') = \underbrace{\exp\!\left(-\tfrac{1}{2}(s_{11} + s_{22})\right)}_{\text{length-dependent prefactor}} \cdot \underbrace{\exp(s_{12})}_{\text{dot-product term}}.
$$
Under the dot-product kernel approximation (length terms approximately constant, see \cref{app:expandernet_to_dpr}), the kernel is proportional to
$$
\exp(s_{12}) = \exp\!\left(\tfrac{1}{d}\,\vx^\top \mA^\top \mA \vx'\right) = \sum_{\ell=0}^{\infty} \frac{1}{\ell!}\left(\tfrac{1}{d}\,\vx^\top \mA^\top \mA \vx'\right)^\ell.
$$
Comparing with the general form $k(t) = \sum_\ell c_\ell\, t^\ell / \ell!$, we read off $c_\ell = 1$ for all $\ell$.
\section{Feature ODE Rise Time Derivations}
\label{app:clo}

This appendix discusses our methodology for arriving at the effective optimization time for MLPs to learn monomial target functions.
Beginning with the feature ODE, we have
\vspace{-0.5em}
\begin{align*}
    \mathcal{L} &= \frac12\left(1-\tilde{c}_\ell\prod_{i=1}^da_i^{\alpha_i}b\right)^2\\
    \dot{a}_i &= -\frac{\partial \mathcal{L}}{\partial a_i} = \left(1-\tilde{c}_\ell\prod_{i=1}^da_i^{\alpha_i}b\right)\frac{\alpha_i\tilde{c}_\ell\prod_{i=1}^da_i^{\alpha_i-1}b}{a_i}\\
    \dot{b} &= -\frac{\partial \mathcal{L}}{\partial b} = \left(1-\tilde{c}_\ell\prod_{i=1}^da_i^{\alpha_i}b\right)\frac{\tilde{c}_\ell\prod_{i=1}^da_i^{\alpha_i}b}{b}\;.\\
\end{align*}
\vspace{-1.0em}
Noting $\frac{1}{\alpha_i}a_i\dot{a}_i = b\dot{b}$, we find $a_i(t) = \sqrt{\alpha_ib^2(t)+a_i^2(0)}$
Plugging this into the feature ODE, we get
\begin{equation*}
    \dot{b} = \left(1-\tilde{c}_\ell\prod_{i=1}^d\left(\valpha_ib^2(t)+a_i^2(0)\right)^{\frac{\valpha_i}{2}}b\right)\tilde{c}_\ell\prod_{i=1}^d\left(\valpha_ib^2(t)+a_i^2(0)\right)^{\frac{\valpha_i}{2}}\;.
\end{equation*}

This equation is not generally solvable, but we can make a number of simplifications to estimate an effective optimization time.
In a gradient flow regime from small initialization, the network output tends to follow two distinct regimes: near a saddle-point and growth \citep{jacot2021saddle, boursier2022gradient, pesme2023saddle}.
We assume that the number of gradient steps to escape from a saddle-point will dominate over the number it takes to reach the next saddle-point, forcing $\left(1-\tilde{c}_\ell\prod_{i=1}^d\left(\valpha_ib^2(t)+a_i^2(0)\right)^{\frac{\valpha_i}{2}}b\right) \approx 1$, which simplifies the ODE to
\begin{equation*}
    \int_0^{b_{f}}\frac{\text{d}b}{\tilde{c}_\ell\prod_{i=1}^d\left(\valpha_ib^2(t)+a_i^2(0)\right)^{\frac{\valpha_i}{2}}} = \int_0^\tau\text{d}t \,,
\end{equation*}
where
\begin{equation}
    b_{f} \equiv \tilde{c}^{\frac{1}{\sum_i \valpha_i+1}}
    \label{eq:b_final}
\end{equation}
represents the final value of $b$ and $\tau$ represents the learning-rate scaled number of gradient steps.
We provide exact results of this integral for the cases of $\valpha_i = 0\; \forall \; i$ and $\valpha_i = 1,\;\valpha_j=0 \; \forall \; j\neq i$ (a constant and linear target).
For higher order integrals, exact results exist, although to high precision we can replace the upper bound $b_{eq}$ with infinity, remaining up to a $\mathcal{O}\left(c^{-\frac{2}{\sum_i\valpha_i+1}}\right)$ constant.
\begin{align}
    \tau_{const} &= \int_0^{\frac{1}{\tilde{c}}} \frac{\text{d}b}{\tilde{c}} = \frac1{\tilde{c}^2}\\
    \tau_{lin} &= \int_0^{\frac{1}{\sqrt{\tilde{c}}}} \frac{\text{d}b}{\tilde{c}\sqrt{b^2+a_i^2(0)}} = \frac{1}{\tilde{c}}\ln\left(\frac{\frac{1}{\sqrt{\tilde{c}}}+\sqrt{\frac{1}{\tilde{c}}+a_i^2(0)}}{a_i(0)}\right) \approx \frac{1}{\tilde{c}}\ln\left(\frac{2}{\sqrt{\tilde{c}}a_i(0)}\right)\\
    \tau_{quad+} &= \int_0^{\tilde{c}^{-\frac{1}{\left(\sum_i\alpha_i + 1\right)}}} \frac{\text{d}b}{\tilde{c}_\ell\prod_{i=1}^d\left(\alpha_ib^2(t)+a_i^2(0)\right)^{\frac{\alpha_i}{2}}}\\
     & = \int_0^\infty \frac{\text{d}b}{\tilde{c}_\ell\prod_{i=1}^d\left(\alpha_ib^2(t)+a_i^2(0)\right)^{\frac{\alpha_i}{2}}}- \mathcal{O}\left(\tilde{c}^{-\frac{2}{\sum_i\alpha_i+1}}\right)
\end{align}
Below, we provide a table of the rise times found with the $\infty$ upper bound.
For shorthand, any $\alpha_i$ that is unspecified takes a value of 0, and $\alpha_{i,j,k,\cdots} = a$ denotes that $\alpha_i = \alpha_j = \alpha_k = \cdots = a$.
Some integrals (ie $\alpha_i=3$, $\alpha_j=1$) are incomplete elliptics---in such cases, we provide a numerical estimate.
We drop the subscript of $\tilde{c}$ as it should be clear from the context.

\begin{itemize}
    \item \textbf{Quadratics} $\begin{cases} 
        \alpha_i = 2 & \tau = \frac{a_0^{-1}}{\tilde{c}} \frac{\pi}{2^{3/2}} \\
        \alpha_{i,j} = 1 & \tau = \frac{a_0^{-1}}{\tilde{c}} \frac{\pi}{2} 
    \end{cases}$

    \item \textbf{Cubics} $\begin{cases} 
        \alpha_i = 3 & \tau = \frac{a_0^{-2}}{\tilde{c}} \frac{1}{\sqrt{3}} \\
        \alpha_i = 2, \alpha_j = 1 & \tau = \frac{a_0^{-2}}{\tilde{c}} \frac{\pi}{4} \\
        \alpha_{i,j,k} = 1 & \tau = \frac{a_0^{-2}}{\tilde{c}} (1)
    \end{cases}$

    \item \textbf{Quartics} $\begin{cases} 
        \alpha_i = 4 & \tau = \frac{a_0^{-3}}{\tilde{c}} \frac{\pi}{8} \\
        \alpha_i = 3, \alpha_j = 1 & \tau = \frac{a_0^{-3}}{\tilde{c}} (0.4623) \\
        \alpha_{i,j} = 2 & \tau = \frac{a_0^{-3}}{\tilde{c}} \frac{\pi\sqrt{2}}{8} \\
        \alpha_i = 2, \alpha_{j,k} = 1 & \tau = \frac{a_0^{-3}}{\tilde{c}} \frac{\pi}{2}(\sqrt{2}-1) \\
        \alpha_{i,j,k,l} = 1 & \tau = \frac{a_0^{-3}}{\tilde{c}} \frac{\pi}{4}
    \end{cases}$

    \item \textbf{Quintics} $\begin{cases} 
        \alpha_i = 5 & \tau = \frac{a_0^{-4}}{\tilde{c}} \frac{2}{3\sqrt{5}} \\
        \alpha_i = 4, \alpha_j = 1 & \tau = \frac{a_0^{-4}}{\tilde{c}} \left( \frac{\pi\sqrt{3}}{27} + \frac{1}{6} \right) \\
        \alpha_i = 3, \alpha_j = 2 & \tau = \frac{a_0^{-4}}{\tilde{c}} (\sqrt{3} - \ln(2 + \sqrt{3})) \\
        \alpha_i = 3, \alpha_{j,k} = 1 & \tau = \frac{a_0^{-4}}{\tilde{c}} \left( \frac{\sqrt{3}}{2} - \frac{\sqrt{2}}{8}\ln(5 + 2\sqrt{6}) \right) \\
        \alpha_{i,j} = 2, \alpha_k = 1 & \tau = \frac{a_0^{-4}}{\tilde{c}} (0.5) \\
        \alpha_i = 2, \alpha_{j,k,l} = 1 & \tau = \frac{a_0^{-4}}{\tilde{c}} (\frac{\pi}{2} - 1) \\
        \alpha_{i,j,k,l,m} = 1 & \tau = \frac{a_0^{-4}}{\tilde{c}} \frac{2}{3}
    \end{cases}$

    \item \textbf{Sextics} $\begin{cases} 
        \alpha_i = 6 & \tau = \frac{a_0^{-5}}{\tilde{c}} \frac{3\pi}{16\sqrt{6}} \\
        \alpha_i = 4, \alpha_j = 2 & \tau = \frac{a_0^{-5}}{\tilde{c}} \frac{\pi(\sqrt{2}-1)}{4} \\
        \alpha_i = 4, \alpha_{j,k} = 1 & \tau = \frac{a_0^{-5}}{\tilde{c}} \frac{\pi}{9} \\
        \alpha_{i,j} = 3 & \tau = \frac{a_0^{-5}}{\tilde{c}} \frac{3\pi}{16\sqrt{3}} \\
        \alpha_{i,j,k} = 2 & \tau = \frac{a_0^{-5}}{\tilde{c}} \frac{3\pi}{16\sqrt{2}} \\
        \alpha_{i,j} = 2, \alpha_{k,l} = 1 & \tau = \frac{a_0^{-5}}{\tilde{c}} \frac{\pi(2-\sqrt{2})}{4} \\
        \alpha_i = 2, \alpha_{j,k,l,m} = 1 & \tau = \frac{a_0^{-5}}{\tilde{c}} \pi(\sqrt{2}-1.25) \\
        \alpha_{i,j,k,l,m,n} = 1 & \tau = \frac{a_0^{-5}}{\tilde{c}} \frac{3\pi}{16}
    \end{cases}$
\end{itemize}
\section{Neuron specialization and target functions}
\label{app:neuron_spec}

We diagnose neuron specialization with a three-panel comparison: training loss, per-direction $\mA_{ii}$ or $\sqrt{\textbf{NFM}_{ii}}$ trajectories, and a heatmap of per-neuron preactivation variance (all $N = 8000$ neurons ranked by variance, plotted against effective optimization time).
The heatmap is the workhorse. Neurons all contribute to the dynamics when there are consistent color changes across neurons at a given time step and there are smooth color gradients. Neuron subpopulations drive the dynamics when groups of neurons change color independently of the rest.
All experiments in this appendix use a biasless network with the cosine activation $\sigma(z) = \sqrt{2}\cos(z + \pi/4)$.

\begin{figure}[H]
    \centering
    \includegraphics[width=\textwidth]{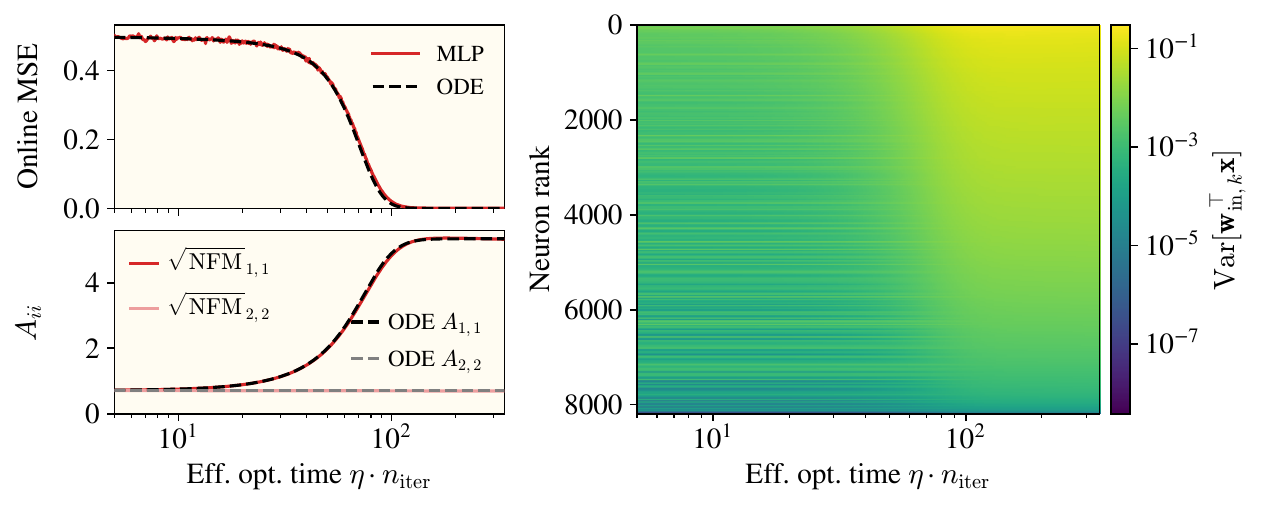}\\
    \includegraphics[width=\textwidth]{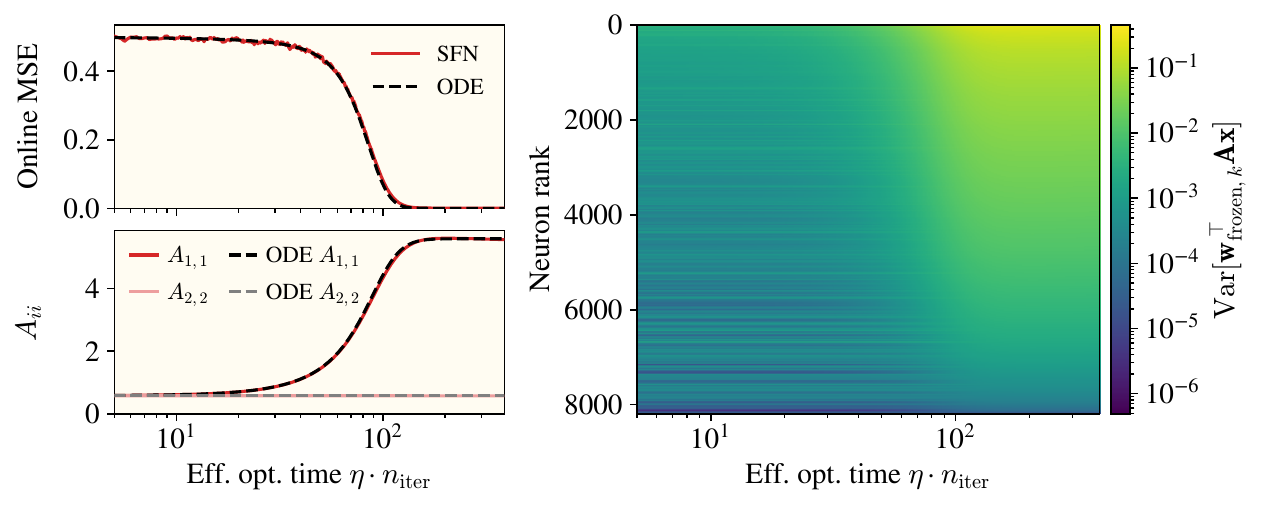}
    \caption{\textbf{Linear target $f^*(\vz) = z_1$.} \emph{Top:} MLP. \emph{Bottom:} SFN. Neurons are sorted by preactivation variance at the end of training in the heatmaps. Loss, $\sqrt{\text{NFM}_{ii}}$ (MLP) or $A_{ii}$ (SFN), and per-neuron preactivation variance. Both models track the feature ODE, and every neuron gains variance on the same timescale.}
    \label{fig:dyn_linear}
\end{figure}

\begin{figure}[H]
    \centering
    \includegraphics[width=\textwidth]{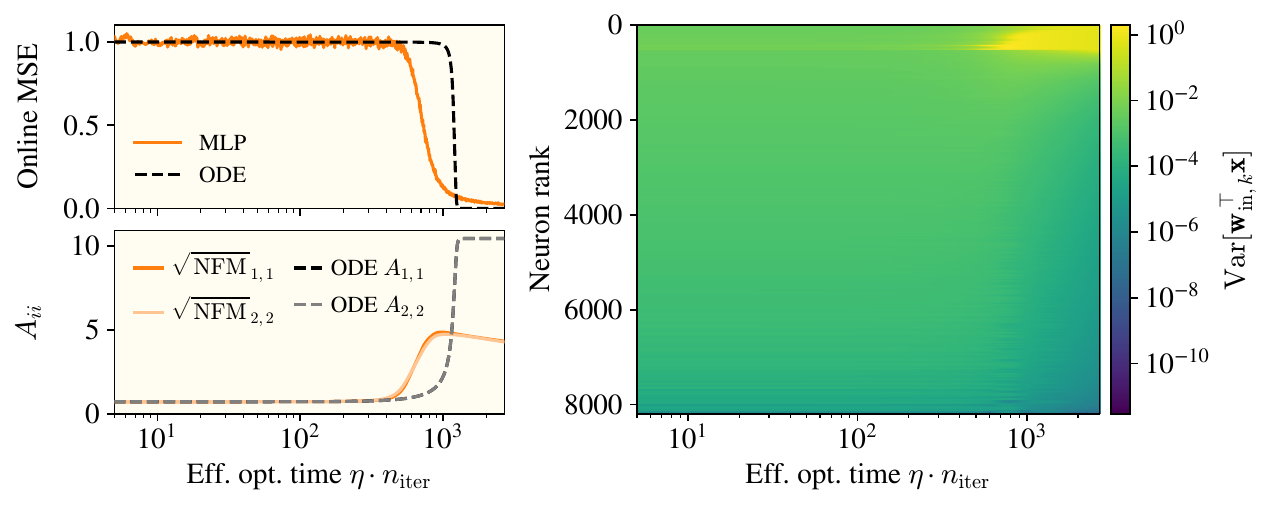}\\
    \includegraphics[width=\textwidth]{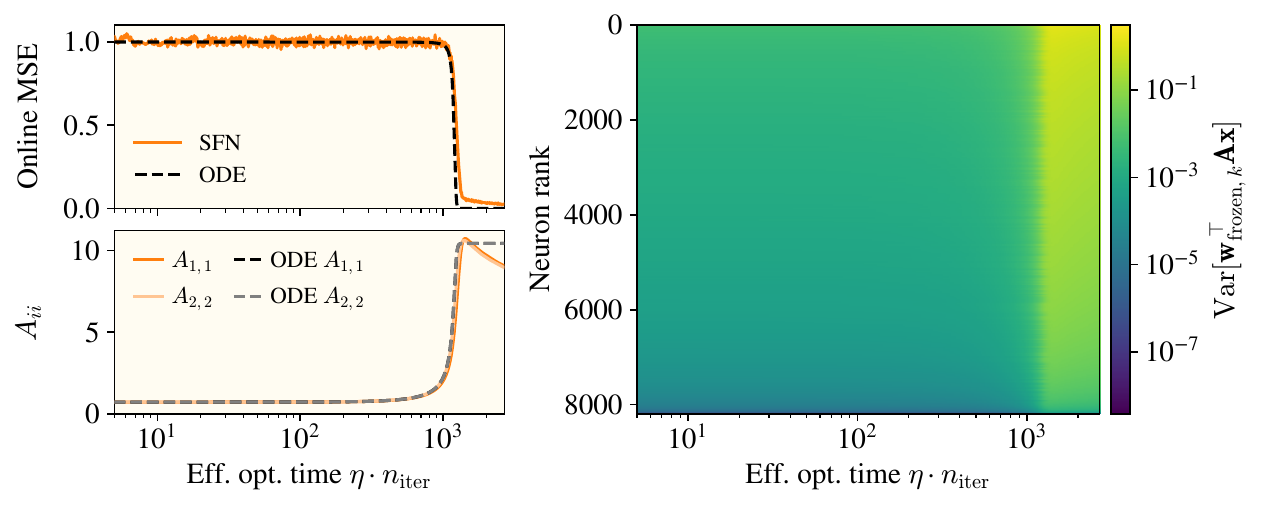}
    \caption{\textbf{Pure-Hermite target $f^*(\vz) = \mathrm{He}_2(z_1) + \mathrm{He}_2(z_2)$.} \emph{Top:} MLP. \emph{Bottom:} SFN. Neurons are sorted by preactivation variance at the end of training in the heatmaps. A specialized subpopulation of roughly 500 MLP neurons (top, right column) gains preactivation variance orders of magnitude above the bulk, drives the loss down earlier than the feature ODE predicts, and pins $\sqrt{\text{NFM}_{ii}}$ below the ODE-predicted $A_{ii}$. The SFN carries no such subpopulation and tracks the ODE.}
    \label{fig:dyn_xnorm}
\end{figure}

\begin{figure}[H]
    \centering
    \includegraphics[width=\textwidth]{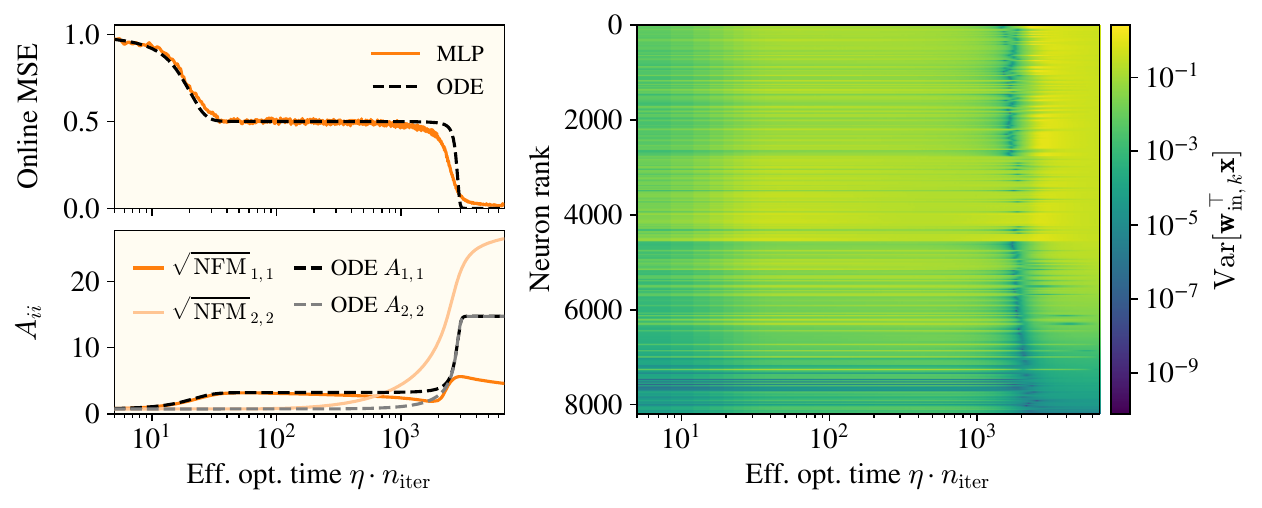}\\
    \includegraphics[width=\textwidth]{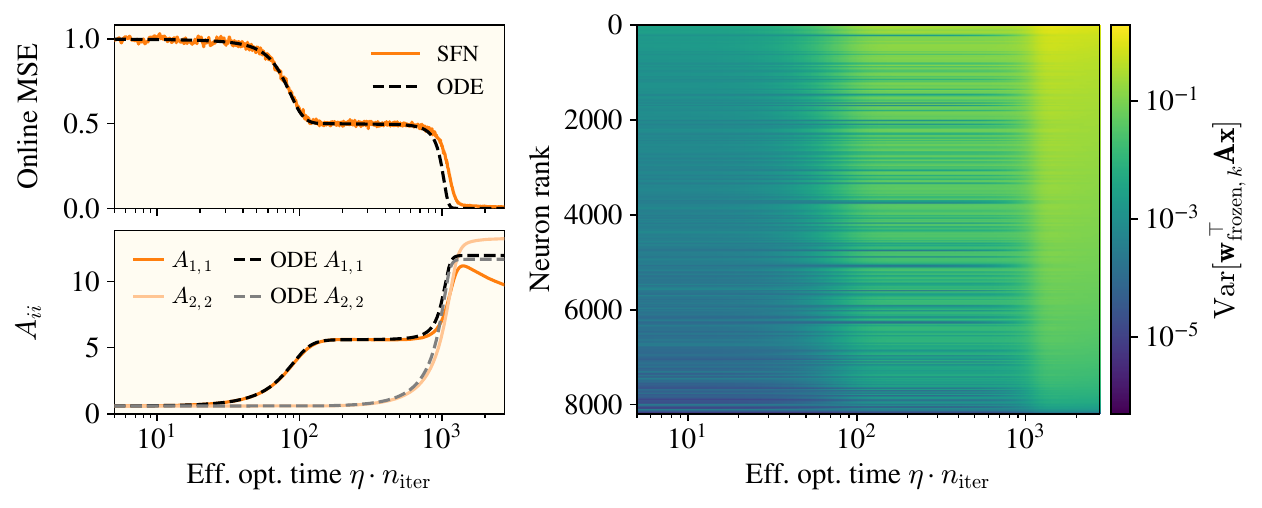}
    \caption{\textbf{Two-term staircase $f^*(\vz) = z_1 + z_1 z_2$.} \emph{Top:} MLP. \emph{Bottom:} SFN. Neurons are sorted by preactivation variance at the end of training in the heatmaps. A faint specialization band in the MLP heatmap accompanies a small late-time drift of $\sqrt{\text{NFM}_{2,2}}$ above the ODE prediction. The staircase loss is otherwise well captured. The SFN matches the ODE throughout.}
    \label{fig:dyn_staircase2}
\end{figure}

DPR has no neurons --- its $\mB^{(\ell)}$ tensors act on all features at once --- so specialization cannot arise.
The frozen weights in the SFN make neuron specialization difficult and we see more little specialized behavior in \cref{fig:dyn_linear,fig:dyn_xnorm,fig:dyn_staircase2}.
The feature ODE in turn predicts the loss and per-direction $A_{ii}$ trajectories well.

MLP neurons can specialize.
On tasks like modular arithmetic, for instance, small subpopulations align with Fourier frequencies of the input \citep{kunin2025alternatinggradientflowstheory,nanda2023progressmeasuresgrokkingmechanistic}.
On the simpler targets we study, whether the MLP specializes depends on the target itself.
On a linear target $f^*(\vz) = z_1$ (\cref{fig:dyn_linear}), neither the MLP nor the SFN specializes much: a linear target produces a per-neuron gradient with no direction-dependent amplification, so all neurons evolve symmetrically.
Their heatmaps are rather uniform, and the feature ODE matches loss and $\sqrt{\text{NFM}_{ii}}$ throughout training.
On a pure-Hermite target $f^*(\vz) = \mathrm{He}_2(z_1) + \mathrm{He}_2(z_2)$ (\cref{fig:dyn_xnorm}), the MLP specializes sharply: roughly 500 neurons at the top of the ranking drive their preactivation variance orders of magnitude above the bulk.
We think this specialized subpopulation drives the MLP loss down earlier than the feature ODE predicts and pins $\sqrt{\text{NFM}_{ii}}$ below the ODE-predicted $A_{ii}$.
On a two-term staircase $f^*(\vz) = z_1 + z_1 z_2$ (\cref{fig:dyn_staircase2}), the MLP specializes only mildly --- faint bands rather than a dominant subpopulation.
The feature ODE captures the staircase structure of the loss and features, with only a small late-time drift in $\sqrt{\text{NFM}_{2,2}}$.

We leave a more detailed study hyperparameter and target function's impact on neuron specialization to future work.

\section{DPR population gradient}
\label{app:gradient}

This appendix derives the DPR population gradient in closed form, using Isserlis' theorem (Wick contractions) to evaluate the Gaussian expectations.
The main text (\cref{sec:ode}) states the resulting gradient flow for some specific target functions. Here, we provide the gradient of DPR for any polynomial target function.

Let $\va_m \in \R^d$ denote the $m$-th row of $\mA$ written as a column vector, so that $(\mA\vx)_m = \va_m\T \vx$, and let $\ve_\nu \in \R^d$ denote the $\nu$-th standard basis vector. Working in the normalized coordinates $\vz = \mGamma^{-1/2}\vx$ (so that $\vz \sim \N(0, \mI_d)$ when $\vx \sim \N(0, \mGamma)$), we expand the target polynomially as $f^\star(\vz) = \sum_{j=0}^{p} \mF^{(j)}\, \vz^{\otimes j}$ with $\mF^{(j)}$ a symmetric order-$j$ tensor.

The main text (\cref{sec:ode}) writes the target in Hermite basis, $f^\star(\vz) = \sum_{\valpha} v_\valpha h_\valpha(\vz)$, with coefficients $v_\valpha$ indexed by multi-index. The monomial form used here is equivalent: standard monomial-to-Hermite expansion identifies $\mF^{(j)}$ with a linear combination of $v_\valpha$ at $|\valpha| \le j$. We use the monomial representation because Isserlis' theorem evaluates Gaussian moments of monomial products directly. All other symbols ($\mA$, $\mB^{(\ell)}$, $\va_m$, $\mGamma$, $\gamma_i$, $c_\ell$) follow the main-text conventions.

\subsection{Derivatives of the DPR loss}

We take the DPR model
\begin{equation}
    f_{\text{DPR}}(\vx) = \sum_{\ell=0}^{p} \sqrt{\frac{c_\ell}{\ell!}}\, \mB^{(\ell)} (\mA\vx)^{\otimes \ell},
\end{equation}
and expand its squared loss
$$
\L(\vtheta) = \E{\vx\sim\N(0,\mGamma)}{\bigl(f_{\text{DPR}}(\vx) - f^\star(\vz)\bigr)^2} = \E{}{f_{\text{DPR}}^2} - 2\,\E{}{f_{\text{DPR}}\, f^\star} + \E{}{(f^\star)^2}.
$$
The last term is parameter-free and drops out of the gradient.
The first two split into a model-model piece (which is quadratic in $\mB$) and a model-target cross piece (which is linear in $\mB$); we compute the parameter derivatives of $f_{\text{DPR}}$ before taking expectations.

\paragraph{$\mB$ gradient.}
At entry $(k_1, \ldots, k_\ell)$ of $\mB^{(\ell)}$, the derivative of $f_{\text{DPR}}(\vx)$ is the matching monomial in $\mA\vx$:
\begin{equation}
    \frac{\partial f_{\text{DPR}}(\vx)}{\partial \mB^{(\ell)}_{k_1,\ldots,k_\ell}}
    = \sqrt{\tfrac{c_\ell}{\ell!}}\, (\mA\vx)_{k_1} \cdots (\mA\vx)_{k_\ell}.
\end{equation}
For the constant term ($\ell=0$), $\mB^{(0)}$ is a scalar and $\partial_{\mB^{(0)}} f_{\text{DPR}} = \sqrt{c_0}$.

\paragraph{$\mA$ gradient.}
An increment in $\mA_{\mu\nu}$ shifts $(\mA\vx)_\mu$ by $x_\nu$, so each order-$\ell$ term contributes $\ell$ identical copies, one per tensor slot:
\begin{equation}
    \frac{\partial f_{\text{DPR}}(\vx)}{\partial \mA_{\mu\nu}}
    = x_\nu \sum_{\ell=1}^{p} \ell \sqrt{\tfrac{c_\ell}{\ell!}}\, \mB^{(\ell)}_{\mu,\, b_2, \ldots, b_\ell}\, (\mA\vx)_{b_2} \cdots (\mA\vx)_{b_\ell},
\end{equation}
where repeated $b$-indices are summed.
The constant term ($\ell=0$) does not depend on $\mA$ and contributes nothing.

\subsection{Expectations via Isserlis' theorem}

Each piece of the loss gradient is the expectation of a polynomial in the jointly Gaussian vectors $\vx \sim \N(0, \mGamma)$ and $\vz = \mGamma^{-1/2}\vx$.
Isserlis' theorem states that for jointly Gaussian zero-mean random variables $X_1, \ldots, X_{2n}$,
\begin{equation}
    \E{}{X_1 X_2 \cdots X_{2n}} = \sum_{\text{pairings } \pi} \prod_{(m,n) \in \pi} \E{}{X_m X_n},
\end{equation}
where the sum runs over the $(2n-1)!!$ partitions of $\{1, \ldots, 2n\}$ into unordered pairs; odd-order moments vanish.
Three types of slots appear in our expressions --- transformed slots $(\mA\vx)_k$, raw slots $x_\nu$ (from the $\mA$ derivative), and target slots $z_f$. Assigning each slot a vector $\vu \in \R^d$,
\begin{equation}
    \vu_k = \va_k \;\;\text{(transformed)}, \qquad \vu_\nu = \ve_\nu \;\;\text{(raw)}, \qquad \vu_f = \mGamma^{-1/2}\ve_f \;\;\text{($\vz$-target)},
\end{equation}
every pairwise covariance takes the unified form $\Cov{}{\,\cdot\,,\,\cdot\,} = \vu_m\T \mGamma \vu_n$. This evaluates to $\va_m\T\mGamma\va_n$ between two transformed slots, $\sqrt{\gamma_f}(\va_m)_f$ between a transformed slot and a $\vz$-slot, $\delta_{ff'}$ between two $\vz$-slots, and analogously for the raw slot.

\paragraph{$\mB$ expectations.}
For the model-model piece, the order-$\ell$ contribution at index $(k_1, \ldots, k_\ell)$ is
\begin{align}
    2 \sqrt{\tfrac{c_\ell}{\ell!}}\, &\E{}{(\mA\vx)_{k_1} \cdots (\mA\vx)_{k_\ell}\, f_{\text{DPR}}(\vx)} \notag \\
    &= 2 \sqrt{\tfrac{c_\ell}{\ell!}} \sum_{i=0}^{p} \sqrt{\tfrac{c_i}{i!}}\, \mB^{(i)}_{b_1,\ldots,b_i}
       \sum_{k\text{-}b\text{ pairings}} \prod_{(m,n)} \va_m\T \mGamma \va_n,
\end{align}
where ``$k\text{-}b$ pairings'' enumerates all unordered pair partitions of the $\ell + i$ indices $\{k_1, \ldots, k_\ell, b_1, \ldots, b_i\}$.
The pairing sum vanishes whenever $\ell + i$ is odd.
The model-target cross piece is analogous,
\begin{equation}
    -2 \sqrt{\tfrac{c_\ell}{\ell!}}\, \sum_{j=0}^{p} \mF^{(j)}_{f_1,\ldots,f_j}
        \sum_{k\text{-}f\text{ pairings}} \prod_{(m,n)} \vu_m\T \mGamma \vu_n,
\end{equation}
with $\vu_m = \va_m$ on $k$-slots and $\vu_m = \mGamma^{-1/2}\ve_m$ on $f$-slots ($\vz$-coordinates of the target).

\paragraph{$\mA$ expectations.}
The same Wick reduction applies, except one slot is now ``frozen'' to the raw coordinate $x_\nu = \ve_\nu\T \vx$ from the $\mA$ derivative:
\begin{align}
    \E{}{\nabla_{\mA} \L}_{\mu, \nu}
    &= 2 \sum_{i=1}^{p} i \sqrt{\tfrac{c_i}{i!}}\, \mB^{(i)}_{\mu,\, b_2, \ldots, b_i}
       \biggl[\, \sum_{j=1}^{p} \sqrt{\tfrac{c_j}{j!}}\, \mB^{(j)}_{c_1, \ldots, c_j}
       \sum_{\nu\text{-}b\text{-}c\text{ pairings}} \prod_{(m,n)} \vu_m\T \mGamma \vu_n \notag \\
    &\hspace{2.4cm} {} - \sum_{j=1}^{p} \mF^{(j)}_{f_1, \ldots, f_j}
       \sum_{\nu\text{-}b\text{-}f\text{ pairings}} \prod_{(m,n)} \vu_m\T \mGamma \vu_n \,\biggr],
\end{align}
where $\vu$ takes value $\ve_\nu$ on the $\nu$-slot, $\va_b$ on $b$- and $c$-slots, and $\mGamma^{-1/2}\ve_f$ on $f$-slots.

\subsection{Full population gradient}

The pieces combine into two equivalent expressions for the full population gradient.

\paragraph{Pairing form.}
Collecting the model-model and model-target reductions,
\begin{align}
    \nabla_{\mB^{(\ell)}_{k_1, \ldots, k_\ell}} \L
    &= 2 \sqrt{\tfrac{c_\ell}{\ell!}} \sum_{i=0}^{p} \sqrt{\tfrac{c_i}{i!}}\, \mB^{(i)}_{b_1, \ldots, b_i}
       \sum_{k\text{-}b\text{ pairings}} \prod_{(m,n)} \va_m\T \mGamma \va_n \notag \\
    &\quad {} - 2 \sqrt{\tfrac{c_\ell}{\ell!}} \sum_{j=0}^{p} \mF^{(j)}_{f_1, \ldots, f_j}
       \sum_{k\text{-}f\text{ pairings}} \prod_{(m,n)} \vu_m\T \mGamma \vu_n,
\end{align}
and the $\mA$ gradient is given by the expression above.

\paragraph{Moment-tensor form.}
Define $\mS = \mA \mGamma \mA\T$ and two Gaussian moment tensors:
\begin{align}
    \mM^{(\ell)}_{k_1, \ldots, k_\ell}
    &= \E{}{(\mA\vx)_{k_1} \cdots (\mA\vx)_{k_\ell}}
    = \sum_{\text{pairings}} \prod_{(m,n)} \mS_{m,n}, \\
    \mM^{(r;\, t)}_{s_1, \ldots, s_r;\, q_1, \ldots, q_t}
    &= \E{}{z_{s_1} \cdots z_{s_r}\, (\mA\vx)_{q_1} \cdots (\mA\vx)_{q_t}},
\end{align}
where $\mM^{(r;\,t)}$ contracts $r$ $\vz$-target slots against $t$ transformed slots through Gaussian pairings.
Using $x_\nu = \sqrt{\gamma_\nu}\,z_\nu$, the raw-slot factor in the $\mA$ derivative pulls a $\sqrt{\gamma_\nu}$ outside the moment, leaving pure $\vz$-and-transformed moments. The gradient then takes the compact contraction form
\begin{align}
    \nabla_{\mB^{(\ell)}_{k_1, \ldots, k_\ell}} \L
    &= 2 \sqrt{\tfrac{c_\ell}{\ell!}} \biggl[\, \sum_{i=0}^{p} \sqrt{\tfrac{c_i}{i!}}\, \mB^{(i)}_{b_1, \ldots, b_i}\, \mM^{(\ell + i)}_{k_1, \ldots, k_\ell,\, b_1, \ldots, b_i} \notag \\
    &\hspace{2.0cm} {} - \sum_{j=0}^{p} \mF^{(j)}_{f_1, \ldots, f_j}\, \mM^{(j;\, \ell)}_{f_1, \ldots, f_j;\, k_1, \ldots, k_\ell} \,\biggr], \\[4pt]
    \nabla_{\mA_{\mu, \nu}} \L
    &= 2\sqrt{\gamma_\nu} \sum_{i=1}^{p} i \sqrt{\tfrac{c_i}{i!}}\, \mB^{(i)}_{\mu, b_2, \ldots, b_i} \biggl[\, \sum_{j=1}^{p} \sqrt{\tfrac{c_j}{j!}}\, \mB^{(j)}_{c_1, \ldots, c_j}\, \mM^{(1;\, i + j - 1)}_{\nu;\, b_2, \ldots, b_i,\, c_1, \ldots, c_j} \notag \\
    &\hspace{3.2cm} {} - \sum_{j=1}^{p} \mF^{(j)}_{f_1, \ldots, f_j}\, \mM^{(j+1;\, i-1)}_{\nu, f_1, \ldots, f_j;\, b_2, \ldots, b_i} \,\biggr].
\end{align}
The pairing form is more explicit and easier to truncate to a single order; the moment-tensor form makes the contraction structure transparent.
Both feed directly into the diagonal-$\mA$ and timescale-separation reductions used in \cref{app:dpr_to_ode}.

\end{document}